\newif\ificml

\ificml

\documentclass[10pt,twocolumn,letterpaper]{article}

\usepackage{iccv}
\usepackage{times}
\usepackage{epsfig}
\usepackage{graphicx}
\usepackage{amsmath}
\usepackage{amssymb}

\usepackage[pagebackref=true,breaklinks=true,colorlinks,bookmarks=false]{hyperref}

\iccvfinalcopy 

\def\iccvPaperID{****} 

\else
    \documentclass[preprint]{elsarticle}
    
    \usepackage{hyperref}
    \usepackage{times}
    \usepackage{epsfig}
    \usepackage{graphicx}
    \usepackage{amsmath}
    \usepackage{amssymb}
    \usepackage{placeins}
    \usepackage{float}

    \journal{Computer Vision and Image Understanding}
    
    \usepackage{numcompress}
    
\fi

\newcommand{\fig}[1]{Figure~\ref{fig:#1}}
\newcommand{\sect}[1]{Section~\ref{sect:#1}}
\newcommand{\tab}[1]{Table~\ref{tab:#1}}

\newcommand{\eq}[1]{(\ref{eq:#1})}

\begin{document}

\ificml

    \title{Set2Model Networks: Learning Discriminatively To Learn Generative Models}

    \author{Alexander Vakhitov\\
    {\tt\small a.vakhitov@skoltech.ru}
    \and
    Andrey Kuzmin\\
    {\tt\small a.kuzmin@skolkovotech.ru}
    \and
    Victor Lempitsky\\
    {\tt\small lempitsky@skoltech.ru} \\
    Skolkovo Institute of Science and Technology\\
    Moscow, Russia
    }
    
    \maketitle

\else
    \begin{frontmatter}
    \title{Set2Model Networks: Learning Discriminatively To Learn Generative Models}
    
    \author{Alexander Vakhitov, Andrey Kuzmin, Victor Lempitsky}
    \address{Skolkovo Institute of Science and Technology\\
    Moscow, Russia}
    \fntext[myfootnote]{\{a.vakhitov,a.kuzmin,lempitsky\}@skoltech.ru}

\fi
\begin{abstract}
\ificml
\else
We present a new ``learning-to-learn''-type approach that enables rapid learning of concepts from small-to-medium sized training sets and is primarily designed for web-initialized image retrieval. At the core of our approach is a deep architecture (a \textit{Set2Model network}) that maps sets of examples to simple generative probabilistic models such as Gaussians or mixtures of Gaussians in the space of high-dimensional descriptors. 
The parameters of the embedding into the descriptor space are trained in the end-to-end fashion in the meta-learning stage using a set of training learning problems. The main technical novelty of our approach is the derivation of the backprop process through the mixture model fitting, which makes the likelihood of the resulting models differentiable with respect to the positions of the input descriptors.

While the meta-learning process for a Set2Model network is discriminative, a trained Set2Model network performs generative learning of generative models in the descriptor space, which facilitates learning in the cases when no negative examples are available, and whenever the concept being learned is polysemous or represented by noisy training sets. Among other experiments, we demonstrate that these properties allow Set2Model networks to pick visual concepts from the raw outputs of Internet image search engines better than a set of strong baselines.
\fi
\end{abstract}

\ificml
\else
    \end{frontmatter}
    
\fi

\section{Introduction}

\begin{figure*}[t]
\centering
\includegraphics[width=0.9\textwidth]{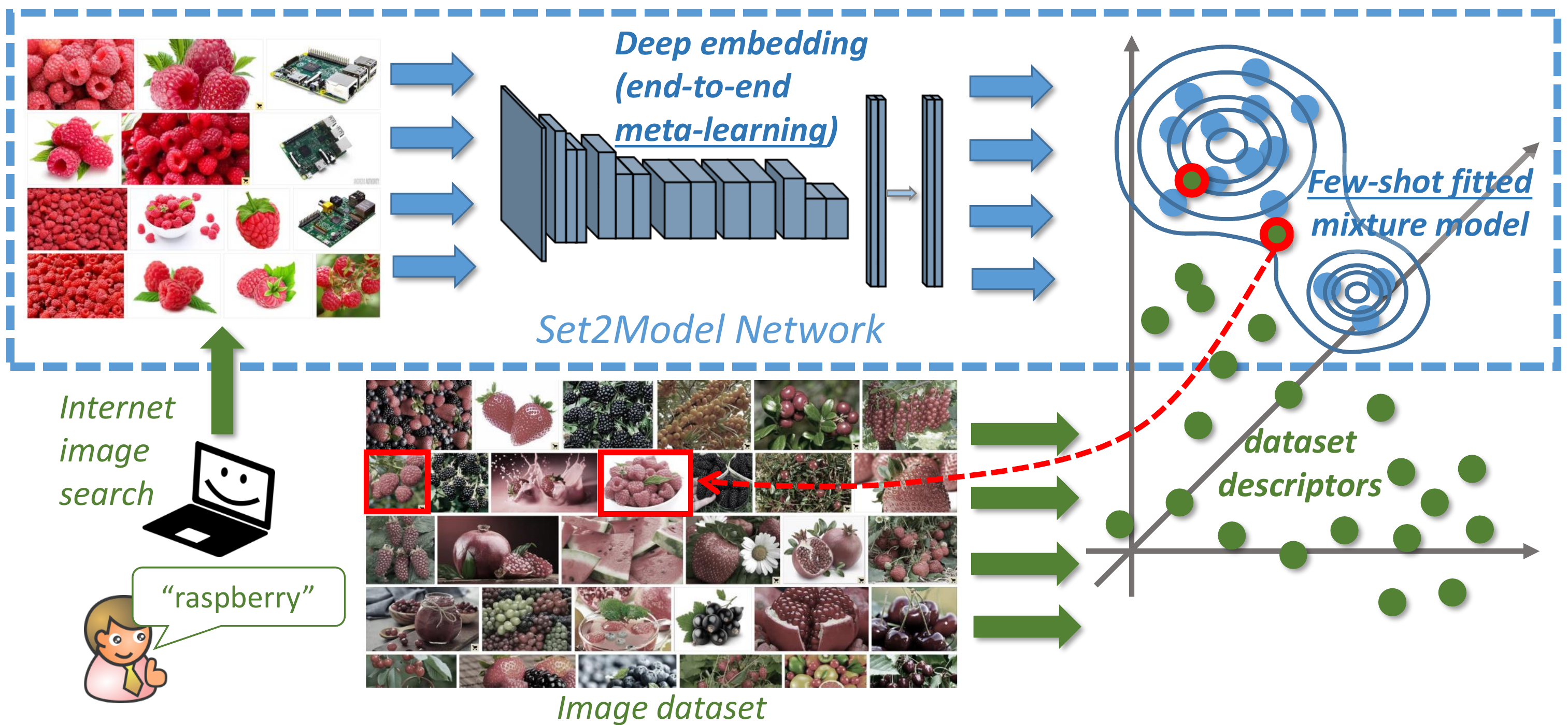}
\caption{\textbf{Top (blue)}: the Set2Model (S2M) network, which takes the set of data points (e.g.\ images), maps them through a non-linear architecture (e.g.\ a deep ConvNet) to a high-dimensional descriptor space, and then fits a generative model (e.g.\ Gaussian mixture) to the resulting set of descriptors. The parameters of the deep embedding are optimized in the end-to-end meta-learning stage, while the generative model is fitted in the few-shot learning stage.
\textbf{Bottom (green)}: our motivating application (Internet-based learning and retrieval). Given a visual concept ``raspberry'', the user obtains a noisy image set depicting raspberries from an Internet image search engine. A pre-meta-learned Set2Model network then maps the set to a mixture model in the descriptor space. Given an unannotated dataset of images, the user can search for images with  raspberries by mapping every image to the descriptor space (using the same deep embedding from the S2M network) and evaluating the likelihood w.r.t.\ the obtained model.}
\label{fig:teaser}
\end{figure*}

The ability to learn concepts from small training sets has emerged as an important frontier in AI. It is well known~\cite{tenenbaum2011grow} that such ability is a hallmark of human intelligence, as humans demonstrate remarkable ability to learn complex visual concepts from only few representative images. Despite surpassing human intelligence in many narrow application domains, AI systems are not able to match this human ability.

Such ability has important practical applications. For example, to pick up a new visual concept, most modern computer vision systems require a set of images depicting this concept. Such an image set is usually mined from the World Wide Web using an Intenet image search engine or comes from a website containing tagged images. Most notable and influential in this respect is the Image-Net project~\cite{deng2009imagenet} that has the goal of obtaining a ``clean'' image set for each of the 80,000+ visual \textit{synsets} corresponding to English nouns. For each such noun, the image set is first obtained by querying the search engine and is then curated through crowd-sourced human labour. The second step (screening by humans) represents a significant burden. As a result, while the positive influence of the Image-Net project on the fields of computer vision and artificial intelligence has been enormous, the progress towards its initial goal has stalled at about one quarter (at the time of submission 21,841 synsets out of 80,000 have been indexed). 

The use of \textit{uncurated} image sets from Internet search engines can potentially enable computers to learn visual concepts automatically and without humans in the loop. Such capability is highly beneficial for intelligent systems, especially in certain scenarios, such as \textit{open-vocabulary} image retrieval that allows users to formulate queries to image collections using arbitrary natural language queries. The use of uncurated image sets obtained from the web, however, is known to be challenging~\cite{Guadarrama14,Kumar14}, since despite the ever-improving performance of image search engines, the returned image sets still contain irrelevant images, since many natural language queries are inherently polysemous, and since many visual concepts often correspond to different visual aspects (e.g.\ outdoor and indoor views of a certain landmark building).

Here, we introduce a new meta-learning approach that is based around a new deep learning architecture called \textit{Set2Model (S2M) network}. An S2M network can be trained to learn new concepts from small-to-medium sized training sets. To map a set of examples to a generative model, a pre-trained S2M network first embeds them into a high-dimensional space, and then fits a generative model in that space to the outputs of the mapping. The training process for the S2M model considers a large set of modeling tasks and corresponds to the tuning of the parameters of the non-linear embedding in order to facilitate easy generative learning for the embedded data.

Unlike analogous meta-learning approaches \cite{tang2010optimizing,vinyals2016matching,wang2016learning} that learn to learn \textit{discriminative} classifiers, a trained Set2Model outputs \textit{generative} models (in this work, we investigate Gaussians and Gaussian mixtures with diagonal covariance matrices). Consequently, a trained Set2Model can learn concepts from only positive examples, which is a more natural setup in many scenarios where negative/background class can be much more diverse than the positive class. Through the use of Gaussian mixtures, S2M networks can also efficiently handle polysemous concepts as well as outliers in the provided training sets. This makes Set2Model networks suitable for picking up concepts from uncurated Internet search engine outputs.

Below, we briefly discuss relevant prior works in \sect{related}, detail our approach in \sect{method}, present results of experimental comparisons in \sect{experiments} and conclude with a short discussion in \sect{conclusion}.

\section{Related work}\label{sect:related}

\paragraph{Meta-learning and Few-shot Learning} Meta-learning (aka ``learning-to-learn'')~\cite{ellis1965transfer,jurgen1996simple} has been a popular approach to handle multi-shot learning scenarios. Interplays between discriminative training and generative probabilistic models in the small training size regime has been investigated in \cite{raina2003classification,Holub05,lasserre2006principled}, where some of the parameters of such models were optimized based on discriminative criteria. Minka~\cite{minka2005discriminative} has pointed out a principled way to derive discriminatively-trainable models. 

Learning concepts from small training sets (sometimes referred to as ``few-shot learning'') has been a subject of intense recent research in meta-learning. The common idea is to learn internal representation, where few-shot learning is simple by observing a large number of few-shot learning tasks. The previous approaches \cite{tang2010optimizing,santoro2016meta,wang2016learning,vinyals2016matching} invariably focused on learning multi-class discriminative classifiers for such few-shot learning problems. Our approach however focuses on single-concept learning problems, which are presented in the form of positive one-class samples. We separate discriminative and generative learning across the two layers of our meta-learning system. In particular, once the embedding within a S2M network is trained discriminatively, a generative learning process in the descriptor space (which is fixed after the S2M training) is used to fit a generative model over those embeddings.

The approach~\cite{Mensink13} learns a metric in the image feature domain in order to improve distance-based image classification and shows that the resulting metric generalizes well to the classes unseen during training. It also proposes a Nearest Class Mean (NCM) classifier as a distance to a mean of image class descriptors which we use as one of the baseline methods.

\ificml
\else
\paragraph{Internet-based learning} Historically, Internet image search relied mostly on textual information surrounding the image using image content to improve ranking. Addressing this, the works \cite{fergus2004visual,Schroff11} focused on how to use visual content in order to improve ranking of the search results. 
A simple bag of visual words model was used in \cite{fergus2004visual,Schroff11}, with \cite{Schroff11} showing that a discrimative SVM classifier performs better than standard information retrieval techniques. 

\cite{Arandjelovic12} addressed large-scale image retrieval problem using image sets as queries and SIFT-based bag-of-words vectors as image features, considering several variants including binary SVM learned on the query set with randomly sampled negatives, ranking using averaged query feature vectors and averaging of the rankings for each individual query from the query set. The works~\cite{Chatfield12,Chatfield15} developed these methods further proposing cascades of classifiers for real-time on-the-fly object category retrieval in large image and video datasets using various features including deep features in \cite{Chatfield15}.


While we consider retrieval from unannotated test collection using a model built from a visual concept query, the work~\cite{Guadarrama14} proposes to generate image-to-text mapping for all the images in the test collection and then to process textual queries using this direct mapping. The approach was evaluated on objects from lab and kitchen environment, as the users were asked to formulate textual queries and the retrieval performance was measured. While this approach performs better than ``blind'' visual matching as in \cite{Arandjelovic12}, it requires computationally intensive preprocessing which in fact is requesting a large database of images with detailed text annotations.

Finally, we mention a number of less related approaches that worked with Internet data. The work~\cite{Kumar14} describes a system for indexing user image collections with the help of Internet search and various datasources such as maps to find locations. In \cite{Tenorth11}, a robotic system is proposed that uses various Internet datasources including image search engine to learn how to perform certain complex activities. Their system learns object classes by training an SVM classifier on image sets resulting from multiple Internet search engine queries. The approach~\cite{Lai10} uses Internet as a source of 3D models to learn 3D object classifiers in point clouds.

Apart from the works from the computer vision and the robotics communities discussed above, our approach is also related to certain directions pursued in the information retrieval and multimedia search communities. These include multiple query retrieval systems (e.g.~\cite{Arandjelovic12}) and a large body of query reranking approaches, some of which use discriminative learning \cite{jain2011learning,yang2010supervised}.

\fi

\section{Set2Model Networks}\label{sect:method}

\newcommand{\T}{\mathcal{T}}

Set2Model networks imply two levels of learning following the "learning-to-learn" principle (e.g., \cite{santoro2016meta}). It implies that a system has two timescales, and the rapid one is associated with learning to solve a task, while the gradual one aims at acquisition of knowledge across tasks. In case of the Set2Model network, on the rapid timescale the network maps a number of samples to a generative model, while on the gradual timescale the parameters of the network are tuned based on a large number of sample class-modeling problems. We call the training process happening at the gradual timescale the {\it meta-learning} stage. The application of the network to a particular problem is called the {\it learning} stage. Below, we discuss the details of these stages, first starting with the learning stage, and then discussing the meta-learning stage. 

\subsection{Learning models with pre-trained S2M networks}

The learning stage (\fig{teaser},top) considers the set of examples $X = \{x^1,x^2,\ldots,x^N\}$, $X \subset {\cal X}$, such as a set of images depicting a certain concept. In the learning stage, the S2M network maps $X$ to a probabilistic model that can be used to evaluate probabilities of other elements belonging to the same concept. 

Modeling the probability distribution in the original space (e.g.\ images) might be overly complex. Therefore, the S2M network firstly maps the elements of $X$ to a specially constructed latent space of descriptors and then uses a simple parametric probability density function (pdf) to model it in the new space. We denote this mapping as $f(x;w):{\cal X} \mapsto \mathbb{R}^n$, whereas $w$ denotes the parameters of the mapping. In this work, we focus on deep convolutional networks as such mappings, though our approach is not specific to a particular architecture.

The mapping $f$ thus transforms the original set $X$ into a descriptor set $D = \{d^1,d^2,\ldots,d^N\}$, where $d^i = f(x^i;w)$. The last stage of the learning process fits a parametric generative model with the pdf $p_{GM}(d; \theta)$ to the set $D$, where $\theta$ are the model parameters. In this work, we consider Gaussian and mixture of Gaussian models with diagonal covariance matrices. We chose a generative approach to descriptor set modelling because it led to better precision while using the pre-learned deep features in our experiments (see Table 1, 'Gauss-PL' vs 'SVM-PL').

The fitting of the model to the set $D$ is performed using maximum likelihood (ML). Thus the parameters $\theta^*$ that maximize the likelihood function $l(\theta|D)$ are sought:
\begin{equation} \label{eq:ml}
\theta^* = \arg\max_{\theta} l(\theta|D),\,\,\, l(\theta|D) = \sum_{i=1}^{N} \log p_{GM}(d^i, \theta).
\end{equation}
Overall, the learning performed by an S2M network can be regarded as the mapping:
\begin{equation}\label{eq:s2m_fun}
F : X \to \theta^*.
\end{equation}
And the relevance of a new data point (e.g.\ an image) $z$ to the concept represented by the set $X$ can be estimated using the obtained density function:
\begin{equation}\label{eq:target_pdf}
p(z|X;w) = p_{GM}(f(z;w), F(X;w)).
\end{equation}
The resulting probabilistic relevance measure can be used e.g.\ to perform retrieval from an untagged set (\fig{teaser},bottom).

\ificml
\else
We also consider adaptive choice of the most appropriate model. In our case this is the choice between a single Gaussian and mixtures with different number of parameters. We use the Bayesian Information Criterion (BIC)~\cite{schwarz1978estimating} to solve the task. This is a general approach proposed for model choice in machine learning when it is impossible to use a validation set. We fit all the available models as described above, compute the value of the criterion:
\begin{equation}\label{eq:BIC}
BIC(k) = 2 k n \log (N) - 2 l(\theta|D),
\end{equation}
where $k$ is the number of mixture components, and choose the model which gives the smallest one.
\fi

\subsection{Meta-Learning S2M Networks}
The goal of the meta-learning is to find the parameters $w$ of the mapping $f(x,w)$ such that the learning process discussed above works well for different concepts.

The meta-learning is performed using the set of tuples $\{\T_i = (X_i, Z_{+,i}, Z_{-,i}) \}$, where each tuple $\T_i$ includes the \textit{concept-describing set} $X_i = \{x_i^1,x_i^2,\ldots,x_i^{N_i}\}$, the \textit{relevant examples set}  $Z_{+,i} = \{z_{+,i}^1, z_{+,i}^2,\ldots,z_{+,i}^{M_{+,i}} \}$ and the \textit{irrelevant examples set} $Z_{-,i} = \{z_{-,i}^1, z_{-,i}^2,\ldots,z_{-,i}^{M_{-,i}} \}$.
For example, $X_1$ can be some set of images of apples from the Internet, the set of $Z_{+,1}$ can be another set of different images also containing apples, and all the images from $Z_{-,1}$ will not contain apples. The second training tuple can then include the sets $X_2$ and $Z_{+,2}$ of pear images and the set of $Z_{-,2}$ of non-pear images, and so on. 

Generally, the meta-learning stage seeks the parameters $w$ such that across all training tuples, the probabilistic relevances estimated using \eq{target_pdf} are higher for the members of the relevant sets than for the members of the irrelevant sets, i.e.:
\begin{equation}\label{eq:rel_ineq}
p(z_{+,i}^k|X_i;w) > p(z_{-,i}^l|X_i;w),
\end{equation}
for various $i,k,l$. 

There are several ways to design loss functions that seek to enforce \eq{rel_ineq}. E.g.\ a loss that draws random elements of relevant and irrelevant sets and computes a monotonic function of the differences in their relevance values. Here, we use a tuple-level loss that directly estimates the probability of the \eq{rel_ineq} to be violated. Given a tuple $(X_i, Z_{+,i}, Z_{-,i})$, let $\theta^*_i$ be $F(X_i;w)$. Using \eq{target_pdf}, we compute the relevance scores for the elements of the relevant and irrelevant sets. 

Let $R_{+,i}$ be the set of relevances for the relevant set $Z_{i,+}$ (i.e.\ $R_{+,i}=\{p(z_{+,i}^1|X_i;w),$ $p(z_{+,i}^2|X_i;w)\dots p(z_{+,i}^{M_{+,i}}|X_i;w) \}$), and let $R_{-,i}$ be the set of relevances for the irrelevant set $Z_{i,-}$. Then the loss based on the probability of the violation of \eq{rel_ineq} can be computed as:
\begin{equation} \label{eq:loss}
L(w) = \sum_i \frac{1}{M_{+,i} M_{-,i}}\sum_{k=1}^{M_{+,i}} \sum_{l=1}^{M_{-,i}} \chi \left[ R_{+,i}^k < R_{-,i}^l \right]\,,
\end{equation}
where $\chi[\cdot]$ returns one if the argument is true and zero otherwise.
Here, each term in the outer summation corresponds to the empirical estimate of the probability of violation of \eq{rel_ineq}.



While the loss \eq{loss} is not piecewise-differentiable, we consider a histogram trick recently suggested in \cite{ustinova2016learning} for metric learning. The idea is to accumulate the relevances for the relevant and irrelevant sets into histograms, and then estimate the required probability (\ref{eq:loss}) using these histograms. Here, to compute the histograms, we fix the triangular kernel density estimator $K(s, \omega)$ that for the argument $s$ and the width parameter $\omega$ is defined as:
\begin{equation}\label{eq:tr_ker}
K(s, \omega) = \max \{1-\frac{2|s|}{\omega},0\}.
\end{equation}
 We choose $l_{\min,i}$ and $l_{\max,i}$ to be the lower and the upper bounds of the numbers in the union of $R_{+,i}$ and $R_{-,i}$, and further accumulate the two normalized histograms $h_{+,i}$ and $h_{-,i}$ spanning the range from $l_{\min,i}$ to $l_{\max,i}$ having $B$ bins each and corresponding to the sets $R_{+,i}$ and $R_{-,i}$ respectively. As discussed in \cite{ustinova2016learning}, the entries of the histograms $h_{+,i}$ and $h_{-,i}$ depend in a piecewise-differentiable manner on the entries of $R_{+,i}$ and $R_{-,i}$.

Given the two histograms, the loss for the tuple $(X_i, Z_{+,i}, Z_{-,i})$ is defined as:
\begin{align} \label{eq:lossAprx}
L(w) = \sum_i \sum_{k=1}^B h_{-,i}^k \sum_{l=1}^B h_{+,i}^l\,,
\end{align}
where $h_{-,i}^k$ and $h_{+,j}^l$ denote the entries of the histograms. Note that the new loss \eq{lossAprx} can be regarded as piecewise-differentiable approximation to the non-differentiable loss \eq{loss}.

Given the loss \eq{lossAprx} (or any other piecewise-differentiable loss enforcing \eq{rel_ineq}), the meta-learning process follows the standard stochastic optimization procedure. The training tuples are sampled randomly, the stochastic approximations of the loss \eq{lossAprx} based on single tuples are computed by forward propagation. During forward propagation, the maximum likelihood fitting \eq{ml} is done by conventional means (e.g.\ closed form for Gaussian distribution, EM-algorithm for Gaussian mixture model). The estimated loss is then backpropagated through the S2M network. Any of the SGD-based optimization algorithms such as ADAM~\cite{kingma2014adam} can be used to update the mapping weights $w$. Backpropagation through the S2M network $F(X,w)$ however relies on the ability to backprop through the maximum-likelihood model fitting \eq{ml}. This backprop step is discussed below.

\subsection{Backpropagation through model fitting}

We now detail the backpropagation through the ML model fitting \eq{ml}, i.e.\ the computation of the partial derivatives $\frac{\partial\theta^*}{\partial d^i_{(j)}}$, where $\cdot_{(j)}$ denotes the $j$-th component of a vector. We start with the Gaussian model, for which this computation is based on a simple closed-form expression, and then proceed to the case of Gaussian mixtures.

In the first case, we consider the Gaussian pdf $p_{G}(d, \theta) = {\cal N}(d, \mu, \Sigma)$:
\begin{equation}\label{eq:gauss}
  {\cal N}(d, \mu, \Sigma) =  (2 \pi)^{-n/2} |\Sigma|^{-1/2} e^{-1/2(d-\mu)^T \Sigma^{-1} (d-\mu)},
\end{equation}
where $\mu$ is a mean and $\Sigma$ is a covariance matrix which we take to be diagonal, and $\theta = \left(\mu^T, \phi^T \right)^T$, denoting $\Sigma = \textrm{diag} \, \phi$.
The optimal (in the ML sense) parameters $\theta^*$ can then be found as:
\begin{equation}
    \mu^* = \frac{1}{N} d^i,     \;\;\; \phi^*_{(i)} = \frac{1}{N}\sum_j \left(d^j_{(i)} - \mu^*_{(i)}\right)^2,
\end{equation}
Differentiation of these formulas w.r.t. the descriptor vectors $d^i$ leads to the following:
\begin{equation}
\nabla_{d^i} \mu^* = \frac{1}{N} {\bf 1}^n,\;\;\;\frac{\partial \phi^*_{(i)}}{\partial d^j_{(k)} } = \delta_{ik} \frac{2}{N} (d^j_{(k)} - \mu^*_{(i)}),
\end{equation}
where ${\bf 1}^n$ is a vector of ones of dimension $n$, $\delta_{ik}$ is a Kronecker symbol, which is zero when $i{\neq}k$ and one for $i{=}k$.

In the case of Gaussian mixtures (GMM), we consider the following pdf:
\begin{equation}\label{eq:gmm}
p_{GMM}(d,\theta) =  \sum_{i=1}^k v_i \, {\cal N}(d, \mu_i, \Sigma_i),
\end{equation}
where $k$ is a number of GMM components, ${\cal N}$ denotes Gaussian pdfs, $\{\mu_i\}_{i=1}^k$ are the means, $\Sigma_i = \textrm{diag}(\phi_i)$ are the diagonal covariance matrices, $\{v_i\}_{i=1}^k$ are weights of the corresponding components and $\theta$ consists of the means, covariance diagonals and weights concatenated. The following constraint on the weights should be satisfied:
\begin{equation}\label{eq:constr}
c(\theta) = \sum_{i=1}^k v_i - 1 = 0.
\end{equation}
As $\theta^*$ delivers maximum to an optimization problem with equality constraints, the method of Lagrange multipliers can be used to derive conditions on the partial derivatives. We consider the scalar multiplier $\lambda^*$ corresponding to the constraint \eq{constr} that maximizes the Lagrangian ${\cal L}(\theta, \lambda)$:
\begin{equation}\label{eq:lagrangian}
{\cal L}(\theta, \lambda) = l(\theta|D) + \lambda c(\theta),\;\;\;(\theta^*,\lambda^*) = \textrm{arg}\max \, {\cal L}(\theta, \lambda),
\end{equation}
The following conditions of optimality then hold for $\theta^*$ and $\lambda^*$:
\begin{equation}\label{eq:lag_theta}
\left\{ \begin{array}{rcl}
\nabla_{\theta} {\cal L}(\theta^*, \lambda^*) & = & 0, \\
\nabla_{\lambda} {\cal L}(\theta^*, \lambda^*) & = & 0.
\end{array} \right.
\end{equation}
Applying differentiation over $d^i$ brings the following system of equations:
\ificml
\else

\begin{equation} \label{eq:syst}
\left\{
\begin{array}{rcl}
\frac{\partial^2}{\partial \theta^2} {\cal L}(\theta^*, \lambda^*) \nabla^T\theta^* + 
\frac{\partial^2}{\partial \lambda \partial \theta} {\cal L}(\theta^*, \lambda^*) \nabla^T \lambda^* + \frac{\partial^2}{\partial d^i \partial \theta} {\cal L}(\theta^*, \lambda^*)  &=& 0,\\
\frac{\partial^2}{\partial \theta \partial \lambda} {\cal L}(\theta^*, \lambda^*) \nabla^T\theta^* + 
\frac{\partial^2}{\partial \lambda^2} {\cal L}(\theta^*, \lambda^*) \nabla^T \lambda^* + \frac{\partial^2}{\partial d^i \partial \lambda} {\cal L}(\theta^*, \lambda^*)  &=& 0,
\end{array} \right.
\end{equation}

\fi
and these are linear equations w.r.t.\ unknown matrix $\nabla \theta^*$ and vector $\nabla \lambda^*$ of partial derivatives w.r.t.\ $d^i_{(j)}$ (in particular, $\nabla \theta^*$ is composed of the values $\frac{\partial\theta^*}{\partial d^i_{(j)}}$ that are sought in this derivation). If $m$ is the dimensionality of $\theta$, then \eq{lag_theta} contains $m+1$ equations. As each of the equation is differentiated by $nN$ variables corresponding to the descriptors, the system \eq{syst} contains $(m+1)nN$ equations on the same number of entries in $\nabla \theta^*$ and $\nabla \lambda^*$. Solving the system \eq{syst} then yields the values of the partial derivatives $\frac{\partial\theta^*}{\partial d^i_{(j)}}$. The linear system solution is performed as the part of the backpropagation process. The system very often becomes sparse allowing for the significant solution process speedup, see appendix A1.


Finally, we note that a similar derivation can be conducted for other generative probabilistic models. Furthermore, one could  address discriminative model fitting (e.g.\ logistic regression) in the same setting as in \cite{tang2010optimizing,santoro2016meta,wang2016learning,vinyals2016matching}, where the set $X$ is augmented with class labels.  

\begin{figure*}
\centering
\includegraphics[width=\textwidth]{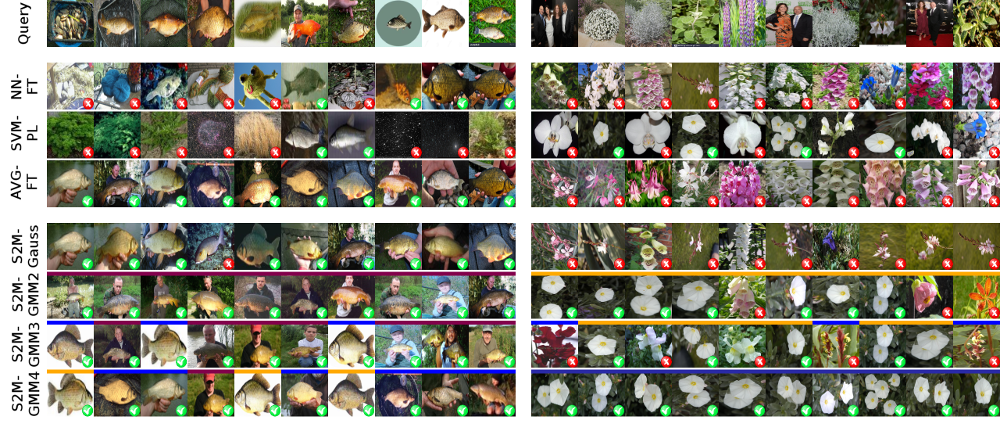}
\caption{Retrieval comparison for the ImageNet dataset (query = `crucian carp,Carassius carassius,Carassius vulgaris,') - left and the Oxford Flowers dataset (query = `silverbush') right. Top row: part of the query image set obtained from Google Search API using the query. Other rows: the top ranked images provided by various methods, namely AVG-FT (average), NN-FT (nearest neighbor), S2M-Gauss (one Gaussian), S2M-GMM2 (mixture of two Gaussians), S2M-GMM3 (mixture of three Gaussians) and S2M-GMM4 (mixture of four Gaussians) using the fine-tuned descriptors, SVM-PL using the pre-learned 'fc8'  features. Color bars encode that image belongs to a certain mixture component. End-to-end-based methods perform better on the given examples. Mixture models successfully filter noisy search engine outputs and capture multiple visual aspects of relevant images.}
\label{fig:joined_output}
\end{figure*}

\section{Experiments}\label{sect:experiments}
Below we provide the experimental evaluation of Set2Model networks (both using single Gaussians and Gaussian mixtures as generative models). We investigate the importance of learning the underlying features. We show that Set2Model networks can be used as generative set models and compare the performance of the Set2Model networks to a number of baselines. Also we apply the Set2Model networks in a few-shot learning problem mainly for the sake of comparison to other meta-learning approaches. 

\ificml
\else
    
    \begin{figure*}
    \centering
    \includegraphics[width=\textwidth]{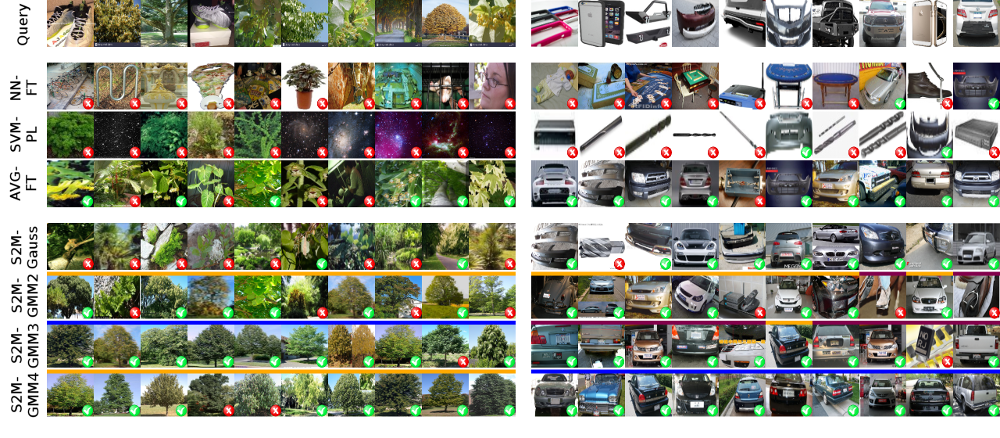}
    \caption{Retrieval comparison for the ImageNet dataset (left: query = `silver lime,silver linden,Tilia tomentosa', right: query=`bumper'). Top row: part of the query image set obtained from Google Search API using the query. Other rows: the top ranked images provided by various methods, namely AVG-FT (average), NN-FT (nearest neighbor), S2M-Gauss (one Gaussian), S2M-GMM2 (mixture of two Gaussians), S2M-GMM3 (mixture of three Gaussians) and S2M-GMM4 (mixture of four Gaussians) using the fine-tuned descriptors, SVM-PL using the pre-learned 'fc8'  features. Color bars encode that image belongs to a certain mixture component. S2M networks often group all relevant query images into a single Gaussian component.}
    \label{fig:flower_imgnet_output}
    \end{figure*}
    
    \begin{figure*}
    \centering
    \includegraphics[width=\textwidth]{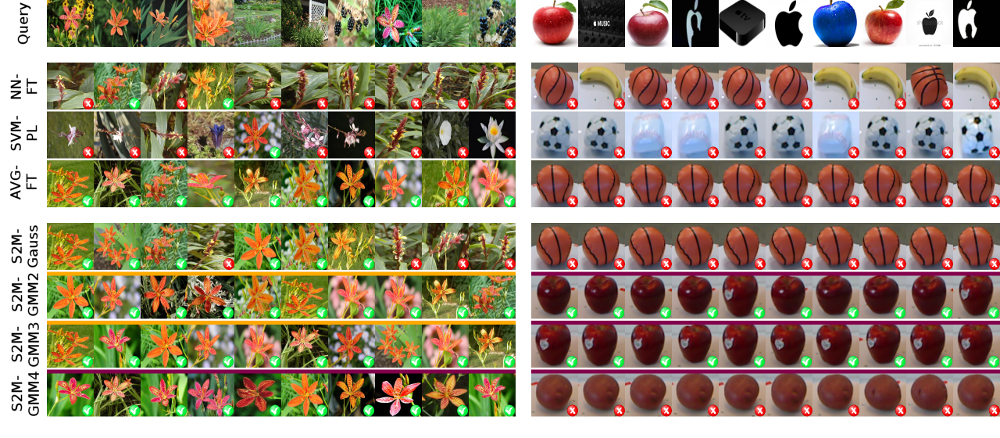}
    \caption{Retrieval comparison for the left and the Oxford Flowers dataset (query=`blackberry Lily') - left and Object RGBD dataset (query = `apple') - right. Top row: part of the query image set obtained from Google Search API using the query. Other rows: the top ranked images provided by various methods, namely AVG-FT (average), NN-FT (nearest neighbor), S2M-Gauss (one Gaussian), S2M-GMM2 (mixture of two Gaussians), S2M-GMM3 (mixture of three Gaussians) and S2M-GMM4 (mixture of four Gaussians) using the fine-tuned descriptors, SVM-PL using the pre-learned 'fc8'  features. Color bars encode that image belongs to a certain mixture component. Highly polysemous queries can be successfully handled by the Set2Model networks. }
    \label{fig:flower_output}
    \end{figure*}

\fi

\subsection{Protocols}

We evaluate the S2M networks in three different sets of experiments. The bulk of the experiments investigates image retrieval with concept-describing image sets generated using Internet image search engines (web-initialized retrieval). Then, we evaluate the possibility of handwritten character retrieval using the Omniglot dataset. Finally, we show that S2M networks can be used for classification of the characters from Omniglot. In particular, to solve the 5-way or 20-way classification problem we build a generative model for each class and compare the ranks w.r.t. these models during test time. The latter experiment shows that the S2M networks can achieve few-shot learning accuracy comparable to the state of the art discriminative approaches although they are not specifically tailored for this task. The results of retrieval experiments are in the \tab{all_map}, and the character classification results are in the \tab{omni_classification}. 

In each case, we split classes into training, validation, and testing. We form the training set out of the training classes, and we train the methods (including ours) on such classes. The methods are compared on the test set. During test, we use the mean average precision (mAP) as the accuracy metric for the retrieval. The meta parameters for all methods are tuned on the validation set.

We perform web-initialized experiments with three different datasets. Google image search is used to obtain the concept-describing sets at all stages (the API for the search engine typically returns 90-100 images). We use class names provided with the datasets to define text queries to the engine. No textual augmentations or search modifiers have been used. Since some of the considered datasets have overlaps with the search engine output, before running the experiments, we identified potential near-duplicates between the Google search results and the datasets using deep descriptors from non-finetuned convolutional network (AlexNet). We then manually checked the potential pairs and removed the true near duplicates from consideration.


\subsection{Implementation details}

We use Caffe~\cite{jia2014caffe} framework to work with convolutional networks. Web-initialized retrieval experiments are based on the AlexNet architecture~\cite{krizhevsky2012imagenet} and character retrieval ones are based on the LeNet architecture~\cite{lecun1998gradient} (Caffe versions are used in all cases). We note that more modern deep convolutional architectures could be used in place of AlexNet or LeNet, however such substitution is likely to benefit all methods equally. For the character classification, we use our Caffe implementation of the network described in  \cite{vinyals2016matching}. It consists of a stack of modules, and each module is a sequence of $3 \times 3$ convolution with 64 filters, batch normalization, Relu non-linearity and $2 \times 2$ max-pooling. It takes $28 \times 28$ images as input and produces 64-dimensional features. 

When performing baseline experiments with pre-learned features, we use 1000-dimensional features that are produced by AlexNet (for the tasks we consider, these features performed optimally or close to optimally compared to other layers). When performing end-to-end learning in case of web-initialized or character retrieval experiments, we replace the last fully connected layer with a smaller one of the same type, of size 128 or 100 respectively.

We perform $l_2$-normalization of the descriptors at the end of the network. In the web-initialized experiments, we start learning from the network weights of the AlexNet provided with Caffe, while the last fully-connected layer for the end-to-end trained architectures is initialized randomly.

The set modelling layer and the loss layer have been implemented using Theano~\cite{bergstra2010theano}, which was used mainly for symbolic differentiation. For back-propagation and learning we use the Caffe implementation of the ADAM algorithm \cite{kingma2014adam} with momentum $0.9$. We choose the learning rate and the termination moment using validation sets. To solve the linear system (\ref{eq:syst}), we utilize its sparsity, since in most cases the coefficient matrices consist of diagonal blocks.

Training of the S2M networks includes an EM algorithm for fitting the GMMs. When a class was encountered for the first time during training, GMM fitting was started with random initialization, and the resulting GMM parameters were memorized. Next time when samples from this class appeared, the saved parameters served as an initial point for the EM algorithm. 

\subsection{Baselines}
Below we describe a set of baseline algorithms. For each of these baselines, we use a certain (not necessarily probabilistic) relevance score $r(z|X;w)$ in the same way as we use $p(z|X;w)$ in the S2M network. The following baselines are considered:
\begin{itemize}
    \item The mean-based system (\textbf{AVG}), which ranks the images in the test set based on the scalar product between their descriptors and the mean of the descriptors $f(x^i,w)$ of the concept-describing set:
    \begin{equation}        
        r_{AVG}(z | X; w) = \left( \frac{1}{N} \sum f(x^i, w) \right)^T z.
    \end{equation}
    
    \item The nearest neighbor (\textbf{NN}) ranker that ranks images in the test set based on the maximum of their scalar product with the query set descriptors:
    \begin{equation}
        r_{NN}(z | X; w) = \max_{i} \left( f(x^i, w) \right)^T z.
    \end{equation}
    \item The support vector machine (\textbf{SVM}) 1-vs-all classifier as in \cite{Arandjelovic12}, where SVM is learned using the query images as positive class and $2|X|$ randomly sampled images from other queries as negative class. If we denote the weight vector of the SVM as $u(\{f(x^i, w)\}_{i=1}^{N})$, then the ranking function can be defined as:
    \begin{equation}
        r_{SVM}(z | X; w) = z^T u(\{f(x^i, w)\}_{i=1}^{N}).
    \end{equation}
\end{itemize}

\ificml
\else

    \begin{table*}[!htbp]
    \small
    \centering
    \begin{tabular}{r|ccccc}
     Model & ImageNet & \begin{tabular}{@{}c@{}}RGBD \\ Object\end{tabular} & \begin{tabular}{@{}c@{}}RGBD \\ Scenes v. 2\end{tabular} & \begin{tabular}{@{}c@{}}Oxford \\ Flowers \end{tabular} & Omniglot \\
    \hline 
    NN-PL  &  0.070 &  0.316  & 0.472  &  0.076 & -  \\
    NN-FT  &  0.073 & 0.318 &  0.490 &   0.528 & 0.145  \\
    SVM-PL & 0.080 & 0.481 & 0.709 & 0.145  & -  \\
    AVG-PL & 0.183 & 0.556  & 0.637  & 0.212 & -\\
    AVG-FT & 0.236 & 0.669  & 0.716  & 0.439 & 0.661\\
    Gauss-PL & 0.186 & 0.529  & 0.625  & 0.240  & - \\
    S2M-Gauss & 0.254& 0.706 & 0.753  & 0.467 & {\bf 0.740} \\
    Gauss-AVG-FT & 0.246 & 0.676 & 0.728  & 0.465 & 0.695 \\
    GMM2-PL & 0.180 & 0.483 & 0.653 & 0.301 & - \\
    S2M-GMM2 & 0.265 & 0.711 & 0.683 & 0.560  & 0.689 \\    
    GMM3-PL & 0.174 & 0.473  & 0.586  & 0.319 & - \\
    S2M-GMM3 & 0.258 & 0.658  & 0.719  & 0.581 & - \\
    GMM4-PL & 0.178 & 0.455 & 0.568 & 0.327 & - \\
    S2M-GMM4 & 0.250 & 0.690 & 0.598 & 0.577 & - \\
    S2M-BGMM-Gauss & 0.257 & 0.667 & {\bf 0.779} & 0.572 & 0.728 \\
    S2M-BGMM-GMM2 & 0.265 & {\bf 0.729} & 0.753 & 0.563 & 0.682\\
    S2M-BGMM-GMM3 & {\bf 0.268} & 0.707 & 0.767 & 0.571 & -\\
    S2M-BGMM-GMM4 & 0.263 & 0.649 & 0.770 & {\bf 0.591} & -\\    
    \end{tabular}
    \caption{Mean average precisions for the experiments (see text for discussion). Top row: dataset. The baselines either use pre-learned ('-PL') or fine-tuned ('-FT') deep features. Gauss-Avg-FT baseline uses features fine-tuned using an AVG baseline (same as AVG-FT), but a Gaussian model on top. Methods with the 'S2M' prefix are the proposed ones. 'S2M-BGMM-XXX' means the S2M network with features fine-tuned as for the 'S2M-XXX' network but using BIC to choose the mixture component number during retrieval. We do not perform any fine-tuning for the SVM, and we do not use GMM with 3 or 4 components with Omniglot due to small size of the concept-describing sets that contain only 10 images. 
    The best achieved results are bolded. The Set2Model networks (S2M-) outperform baselines. End-to-end finetuning improves the results considerably for all methods ('-PL' vs '-FT'). Adaptive model choice using BIC improves precision. Also, using correct end-to-end learning for a single Gaussian (S2M-Gauss) performs better than using end-to-end learning for mean-based retrieval, while using Gaussian models fitted to resulting features during retrieval (Gauss-AVG-FT).}
    \label{tab:all_map}
    \end{table*}        
    
    \begin{table}    
         \centering
        \begin{tabular}{r|c|c}
             Model &  5-way & 20-way \\
             \hline
             Matching Nets \cite{vinyals2016matching}& {\bf 0.989} & {\bf 0.985} \\ 
             MANN (No Conv) \cite{santoro2016meta} & 0.949 & - \\
             Convolutional Siamese Net \cite{koch2015siamese} & 0.984 & 0.965 \\
             S2M-Gauss & 0.985 & 0.956
         \end{tabular}
         \caption{Results for the 5-shot 5-way or 20-way classification on the Omniglot dataset. We performed meta-learning of S2M-Gauss based on the same underlying deep network as described in \cite{vinyals2016matching} using the protocol described in \sect{method}. During classification, we choose the class label corresponding to the maximal rank produced by the S2M network for 5 (or 20) considered classes. The testing protocol follows \cite{vinyals2016matching}.} 
         \label{tab:omni_classification}
    \end{table}    
\fi

We evaluate these baseline methods as well as single Gaussian and GMM models for the pre-learned descriptors('\textbf{-PL}' in \tab{all_map}). We also consider fine-tuning of the convolutional network for the mean~(AVG) and the nearest neighbor~(NN) ranking ('\textbf{-FT}' in \tab{all_map}). In this case, we use exactly the same learning architecture as explained in the previous section, but plug the corresponding relevance measure $r_{NN}(Z | X; w)$ or $r_{AVG}(z | X; w)$ instead of (\ref{eq:target_pdf}) into the computation of the histogram loss. 

Finally, our strongest baseline (\textbf{Gauss-AVG-FT} in \tab{all_map}) is an ablated S2M network that is fine-tuned for the mean-based retrieval, but uses Gaussian model fitting during retrieval in the same way as our model based on single Gaussian does.
\ificml
\else
Alongside the baselines, we report the results of our system (S2M network) for different number of Gaussians in the mixtures: for the scenario when we use exactly same number of mixture components during retrieval and during meta-learning ('\textbf{S2M-Gauss}', '\textbf{S2M-GMM}m', $m=2,3,4$) as well as for the scenario when the number mixture components is fixed during meta-learning but chosen using BIC during retrieval ('\textbf{S2M-BGMM-Gauss}', '\textbf{S2M-BGMM-GMM}m', $m=2,3,4$). We suppose that meta-learning with the model having a particular number of components does not necessarily mean that the same number of components should be used for learning, taking into account that a mixture model with $k_1$ components can be considered a particular case of a mixture model with $k_2>k_1$ components when $k_2-k_1$ weights are exactly zero.

\fi

\subsection{Results}

The quantitative results for all datasets are summarized in Tables \ref{tab:all_map}, \ref{tab:omni_classification}. We also illustrate retrieval performance of some of the compared methods at the 
\ificml
\else
Figures~\ref{fig:joined_output},~\ref{fig:flower_imgnet_output},~\ref{fig:flower_output}. In these figures we indicate whether the image truly belongs to the query class by showing a corresponding symbol in the bottom-right corner of the image. Images from the output of the S2M networks also have a colored bar encoding a particular mixture component 'responsible' for this image. We now discuss the considered datasets and the results on them.
\fi

\paragraph{ImageNet} %
Our first experiment uses classes from the ImageNet dataset \cite{deng2009imagenet}. Since we use networks pretrained on the ILSVRC classes \cite{ILSVRC15}, we made sure that 1000 classes included into the ILSVRC set are excluded from our experiments. 

To perform the experiments, we selected 509 random synsets for training, 99 synsets for validation and 91 synset for testing. The results (\tab{all_map}) demonstrate that end-to-end learning is able to improve the mAP of the baseline methods by 1-4 percent and of the proposed methods based on distribution fitting by 5-7 percent. Importantly, the gap between the model that fits a single Gaussian and the model that uses the mean vector is almost 2 percent. Using mixture models with two or three components improves the performance further.

\begin{figure} 
\centering
\ificml
\else
    \includegraphics[width=\textwidth]{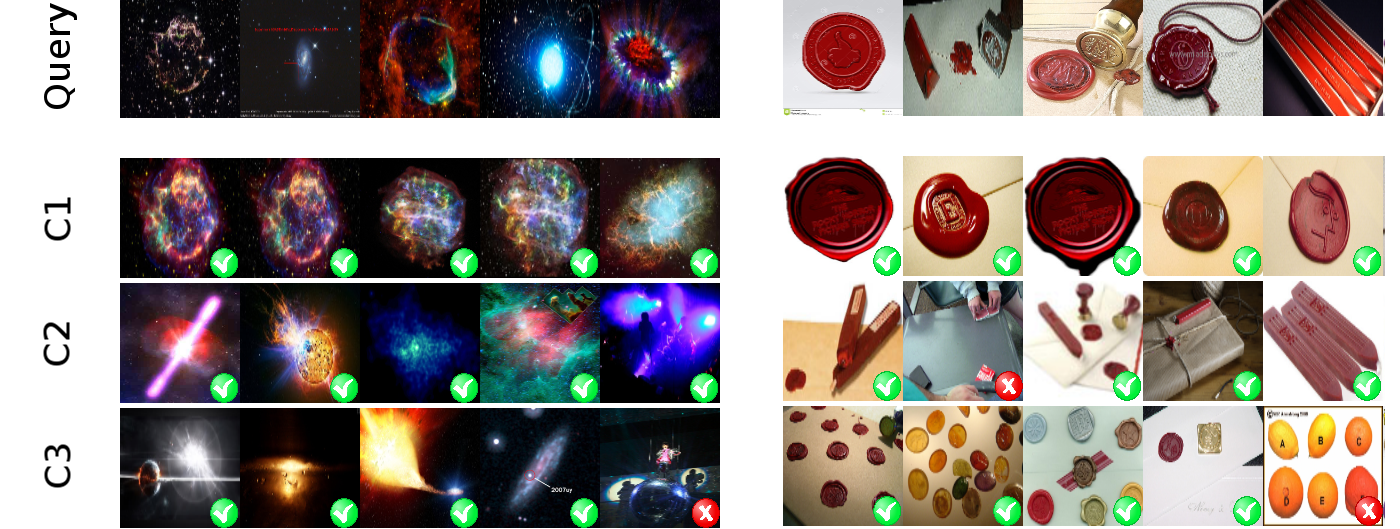}
\fi

\caption{Set2Model networks can handle polysemous queries in a natural way. Top: part of the query image set obtained from Google Search API~(queries='supernova' and 'seal, sealing wax'), bottom: relevant images retrieved by the three components of the GMM3 model returned by the corresponding S2M network. In these cases, each mixture component captures a certain aspect of a visual concept.}\label{fig:components}
\end{figure}

We also observe, that uncurated Google image search outputs for some of the ImageNet synsets are very noisy, S2M networks often group all relevant query images into a single Gaussian component 
\ificml
\else
    (\fig{joined_output}-right, \fig{flower_imgnet_output}).
\fi
This is often a desirable performance, since being able to absorb irrelevant aspects of the query into a separate component may allow to learn a better model for the relevant aspect. At the same time, when multiple aspects are relevant, multiple mixture components are often able to retrieve them as well (\fig{joined_output}-left and \fig{components}).
\ificml
\else
We show an orthogonal projection of the 128-dimensional descriptors of one of the test classes from this dataset, of another test lasses and of the Google outputs corresponding to the chosen class together with the relevance isolines for a single Gaussian and a GMM onto a plane to
show how a model with multiple Gaussians can more accurately model a distribution than a single Gaussian and how a single mixture component groups the relevant descriptors, while the query can contain other visual concepts as well, see \fig{imgnet_descriptor_viz}. 

The model choice with the BIC criterion according to (\ref{eq:BIC}) gives the best precision for this dataset. It is achieved in the deep domain constructed by a S2M network with three mixture components (see \tab{all_map}).  
\fi

\ificml
\else
\begin{figure}
\centering
\includegraphics[width=\textwidth]{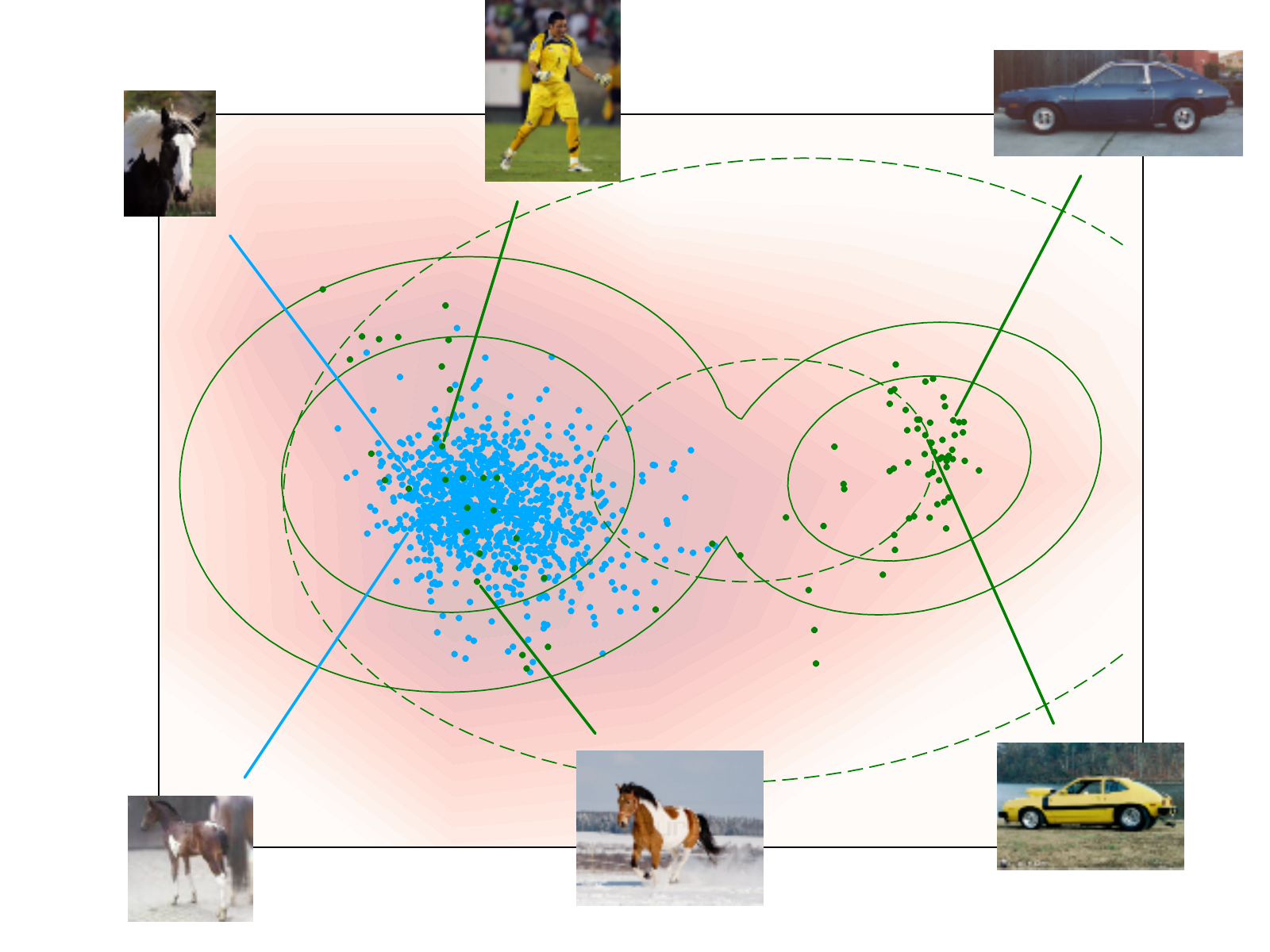}
\caption{Comparison of the Gaussian and mixture of two Gaussians  used by the S2M Network to model the image descriptor distribution. The descriptors of images from the the Google Image Search output (query=`pinto') - green, the corresponding ImageNet category - light blue, other ImageNet categories from the test set - red are orthogonally projected onto a plane with the help of Partial Least Squares Regression. The same approach is used to project isolines of the models for the query image descriptors returned by the Set2Model network (dashed green for Gaussian and solid green for two component Gaussian mixture (GMM2)). The descriptors are computed using the end-to-end learning of a Set2Model network with GMM2 model. The descriptor distribution is represented by GMM2 more accurately than by a single Gaussian, each mixture component captures a separate visual entity of a polysemous query (horse vs car), and only one mixture component corresponds to the visual concept relevant for the ImageNet images (horse).}
\label{fig:imgnet_descriptor_viz}
\end{figure}
\fi

\paragraph{RGBD} This experiment uses the RGBD-Object~\cite{lai2011large} 
\ificml
\else
    and RGBD-Scenes v.2~\cite{lai2014unsupervised} datasets (the latter is used solely for testing). 
\fi
RGBD-Object contains multiple view images of 200 tabletop objects of 51 category
\ificml
\else
    , and for the particular 5 of them the RGBD-Scenes~v.2 contains the RGBD sequences of indoor scenes where they are present, from which we have segmented the object images.
\fi

The amount of relevant images in the Internet search results in this case differs greatly from category to category. For the polysemous categories such as 'apple', the proposed generative models can give significant benefit 
\ificml
\else
(\fig{flower_output}-right). 
\fi
The results in \tab{all_map} demonstrate even greater improvements from the end-to-end learning (perhaps due to a  bigger domain gap between the ImageNet and this dataset). Such training improves the mAP for the mean classifier by nine percent and for the proposed methods by 11-14 percent. The methods based on distribution fitting again perform better. 
\ificml
\else
Due to domain shift, S2M Network demonstrates smaller gains when tested on the Scenes v.2 dataset. 
\fi
\ificml
\else
    The best precisions are again achieved by the S2M network with model choice based on the BIC criterion, illustrating the fact that this approach is more flexible and reflects the diverse complexity of concept-describing sets (e.g.\ a polysemous and noisy query 'apple' and a query 'banana' with a very precisely defined visual concept).
\fi




\paragraph{Oxford Flowers}
We use the Oxford Flowers-102 dataset \cite{nilsback2008automated}, consisting of images of 102 different UK flowers. The dataset was split into 80 categories for training and validation, and 22 for testing. The results in the \tab{all_map} show that end-to-end learning procedure improves the mAP on 14 percent for the mean classifier, 16-19 percent for the proposed models. \fig{joined_output} shows an example of the polysemous and noisy query ('silverbush'), where the ability of the Gaussian mixture models to capture multi-modal distributions provides our approach a big advantage. 
\ificml
\else
\fig{flower_output}-left shows an output for the query 'blackberry Lily' where even simple descriptor averaging can bring robustness to the system due to the fact that searched collection is much less diverse than the query. Model choice with the help of the BIC criterion achieves the best mAP again. It illustrates flexibility of this approach. Note that meta-learning with larger number of components works better on this dataset, since the second-best result belongs to a Gaussian mixture of three components.
\fi

\ificml
\else
\begin{figure}
\centering
\includegraphics[width=\textwidth,trim={1cm 1cm 1cm 1cm},clip]{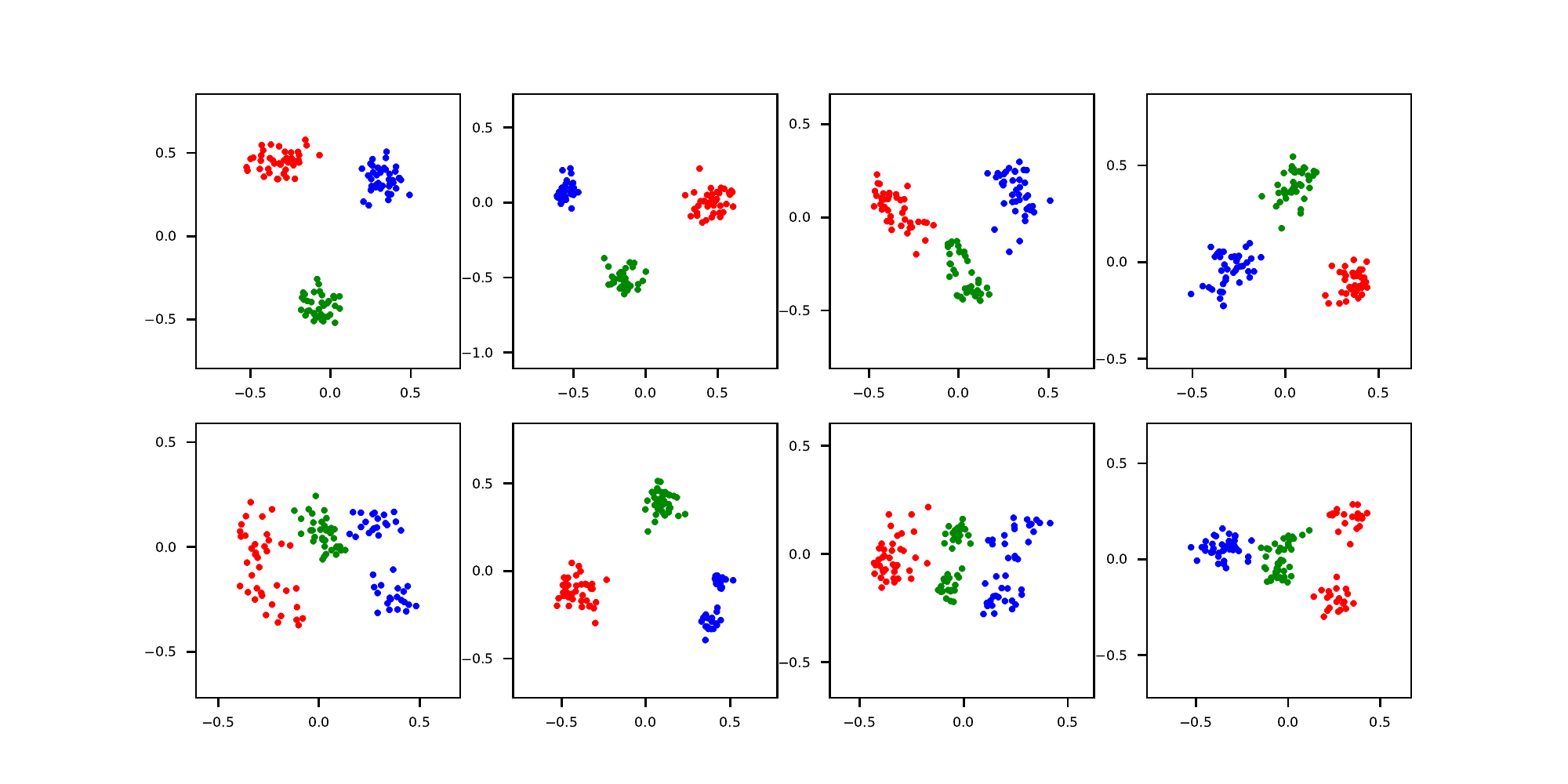}
\caption{Comparison of the learned descriptor distributions for averaging (top) and Gaussian (bottom) on Omniglot dataset, projected using Partial Least Squares Regression. For each column we have chosen triplets of training classes, represented by a separate color. Gaussian-based modeling allows for a more flexible model in the descriptor space, which is reflected in better accuracy of retrieval (\tab{all_map}).}
\label{fig:omni_viz}
\end{figure}
\fi
\paragraph{Omniglot}
Finally, we used the Omniglot dataset~\cite{lake2011one} (which has become the standard testbed for meta-learning methods) to test the ability of the S2M network to use small concept-describing sets for generative model construction. The Omniglot dataset consists of 20 hand-drawn images for each of the 1623 characters from different alphabets. In the retrieval experiment, for learning and testing we use concept-describing, relevant and irrelevant sets of ten images, composing one batch of five training tuples $(X_i, Z_{+,i}, Z_{-,i})$. We randomly split the dataset into 1200 classes for validation and training, and 423 for testing. We rotate images by randomly generated multiples of 90 degrees during testing and training, following  \cite{vinyals2016matching}. Due to the small size of the concept-describing sets, S2M networks based on mixtures of two Gaussians are less accurate, and mixtures of more components were not evaluated. The results in \tab{all_map} show that single Gaussian model performs best
\ificml
\else 
, and \fig{omni_viz} illustrates that Gaussian model handles more diverse descriptor distributions than the averaging baseline allowing for more complex descriptor distributions to emerge. 
\fi

Furthermore, to compare against the recent few-shot learning methods we performed a classification experiment on this dataset. During meta-learning, we used a batch consisting of three training tuples with $|X_i| = 5$, $|Z_{+,i}|=15$, $|Z_{-,i}|=20$. During test time, to do $c$-way classification we build $c$ models using the S2M network and choose a class label corresponding to the model producing a maximal rank for the test example. We compare the results to the state of the art methods at the \tab{omni_classification}. Although the S2M network is not trained to discriminate between classes, still it can be used this way during test time and provide competitive results: better than \cite{santoro2016meta}, similar to \cite{koch2015siamese} and not much worse than \cite{vinyals2016matching}.


\section{Summary} \label{sect:conclusion}

In this work we have proposed Set2Model networks as a new architecture for meta-learning that is particularly suitable for retrieval applications, where queries are given as sets of positive samples. The Set2Model networks are able to map such queries into probabilistic models in specially-designed descriptor spaces. The parameters of such descriptor embeddings are optimized end-to-end, while taking the model fitting into account. We have shown experimentally that such proper end-to-end training is beneficial for the retrieval quality.

In order to gain the ability to handle mixture model fitting within our approach, we have derived a way for backpropagation through the maximum likelihood model fitting. We have presented a number of experiments for image retrieval based on noisy image sets obtained from the Internet image search engines as well as for the hand-drawn character retrieval that show the ability of the S2M networks to generalize across classes and handle the challenges of visual concept modeling from small and medium-sized training sets better than baseline models. We have also shown that generatively trained S2M networks can achieve similar accuracy to the state of the art in few shot learning problems.



\ificml
\else
\section{A1. Typical structure of a system matrix for implicit function differentiation} \label{gmm_diff_formulas}
In the following appendix, we derive formulas for the required derivatives of the likelihood function for the Gaussian Mixture models with diagonal covariance matrices, and show that very often the second derivative matrices become sparse during meta-learning, which accelerates the process significantly.

The log-likelihood function $l(\theta|D)$ is a sum of one-observation likelihood functions as given in 
\ificml
the main text, eq.~(1).
\else
(\ref{eq:ml}).
\fi

{\bf Theorem.} If log-likelihood function $\log h(\theta|d)$  corresponding to a mixture model can be represented as a logarithm of a sum of multiplied coordinate-wise likelihood functions with non-intersecting parameter sets:
\begin{equation}
\log h(\theta|d) = \log \{ \sum_{i=1}^k v_i \prod_{j=1}^n g(\theta_{i,j} | d)\},
\end{equation}
where $v_i$ are mixture weights, $v_i \geq 0$, $\sum_{i=1}^k v_i = 1$, $\theta = [\theta_1^T,\theta_2^T,\ldots,\theta_k^T, v_1,v_2,\ldots,v_k]^T$ is the vector of the model parameters, $\theta_i = [\theta_{i,1}^T,\theta_{i,2}^T,\ldots,\theta_{i,q}^T]^T$ is the vector of one mixture component's parameters, $\theta_{ij} = [\theta_{i,j,1}, \theta_{i,j,2}, \ldots,\theta_{i,j,n_c}]^T$ is the vector of the  coordinate-wise likelihood function parameters of length $n_c$, $g(\theta_{i,j} | d)$ is a coordinate-wise likelihood function for the coordinate $j$, {\em then} the second derivatives of $\log h(\theta_d)$ w.r.t. the parameters of the model (except the weights) are
\ificml

    \begin{equation*}
        \frac{\partial^2 \log h(\theta|d)}{\partial \theta_{i,j,k} \partial \theta_{u,s,t}}  = 
    \end{equation*}
    
    \begin{equation}
    \left\{
    \begin{array}{lr}
    (r_i - r_i^2) \frac{g'_{k}(\theta_{i,j} | d_{(j)})g'_{t}(\theta_{i,s}|d_{(s)})}{g(\theta_{i,j} | d_{(j)})g(\theta_{i,s} | d_{(s)})}, & i=u, j \neq s,\\
    r_i \frac{g''_{kt}(\theta_{i,j} | d_{(j)})}{g(\theta_{i,j} | d_{(j)} )} - r_i^2 \frac{g'_{k}(\theta_{i,j} | d_{(j)})g'_{t}(\theta_{i,j} | d_{(j)})}{(g(\theta_{i,j} | d_{(j)}))^2}, & i=u, j=s, \\
    -\frac{1}{h^2(\theta|d)} r_i r_u
    \frac{g'_{k}(\theta_{i,j}|d_{(j)})}{g(\theta_{i,j} | d_{(j)})}
    \frac{g'_{t}(\theta_{u,s}|d_{(s)})}{g(\theta_{u,s} | d_{(s)})}, &
    i \neq u,
    \end{array} \right.
    \end{equation}
    
\else

\begin{equation}
\frac{\partial^2 \log h(\theta|d)}{\partial \theta_{i,j,k} \partial \theta_{u,s,t}}  = 
\left\{
\begin{array}{lr}
(r_i - r_i^2) \frac{g'_{k}(\theta_{i,j} | d_{(j)})g'_{t}(\theta_{i,s}|d_{(s)})}{g(\theta_{i,j} | d_{(j)})g(\theta_{i,s} | d_{(s)})}, & i=u, j \neq s,\\
r_i \frac{g''_{kt}(\theta_{i,j} | d_{(j)})}{g(\theta_{i,j} | d_{(j)} )} - r_i^2 \frac{g'_{k}(\theta_{i,j} | d_{(j)})g'_{t}(\theta_{i,j} | d_{(j)})}{(g(\theta_{i,j} | d_{(j)}))^2}, & i=u, j=s, \\
-\frac{1}{h^2(\theta|d)} r_i r_u
\frac{g'_{k}(\theta_{i,j}|d_{(j)})}{g(\theta_{i,j} | d_{(j)})}
\frac{g'_{t}(\theta_{u,s}|d_{(s)})}{g(\theta_{u,s} | d_{(s)})}, &
i \neq u,
\end{array} \right.
\end{equation}

\fi
where responsibility of the $i$-th component is defined as a fraction $r_i = \frac{v_i \prod_{j=1}^q g(\theta_{i,j} | d_{(j)})}{h(\theta|d)}.$

{\bf Proof.} 


The first derivative is:
\begin{equation}
\frac{\partial}{\partial \theta_{i,j,k}} \log h(\theta|d) = \frac{1}{h(\theta|d)} v_i \prod_{m=1}^q g(\theta_{i,m} | d_{(m)}) \frac{g'_{k}(\theta_{i,j}|d_{j})}{g(\theta_{i,j} | d_{(j)})}.
\end{equation}
In case when both differentiation variables are the parameters of the same mixture component, the second derivative is:
\ificml
$$
\frac{\partial^2}{\partial \theta_{i,j,k} \partial \theta_{i,s,t}} \log h(\theta|d) = \frac{1}{h(\theta|d)} v_i \prod_{m=1}^q g(\theta_{i,m}|d_{(m)})
$$
\begin{equation}
\frac{g'_{k}(\theta_{i,j}|d_{(j)})g'_{t}(\mu_{i,s}|d_{(s)})}{g(\theta_{i,j}|d_{(j)})g(\theta_{i,s}|d_{(s)})} - 
\end{equation}
\begin{equation}
-\frac{1}{h^2(\theta|d)} v_i^2 \left( \prod_{m=1}^q g(\theta_{i,m}|d_{(m)}) \right)^2 \frac{g'_{k}(\theta_{i,j} | d_{(j)})g'_{t}(\theta_{i,s}|d_{(s)})}{g(\theta_{i,j} | d_{(j)})g(\theta_{i,s} | d_{(s)})} =
\end{equation}
\begin{equation}
=(r_i(d) - r_i^2(d)) \frac{g'_{k}(\theta_{i,j} | d_{(j)})g'_{t}(\theta_{i,s}|d_{(s)})}{g(\theta_{i,j} | d_j)g(\theta_{i,s} | d_{(s)})}.
\end{equation}
\else
\begin{equation}
\frac{\partial^2}{\partial \theta_{i,j,k} \partial \theta_{i,s,t}} \log h(\theta|d) = \frac{1}{h(\theta|d)} v_i \prod_{m=1}^q g(\theta_{i,m}|d_{(m)}) \frac{g'_{k}(\theta_{i,j}|d_{(j)})g'_{t}(\mu_{i,s}|d_{(s)})}{g(\theta_{i,j}|d_{(j)})g(\theta_{i,s}|d_{(s)})} - 
\end{equation}
\begin{equation}
-\frac{1}{h^2(\theta|d)} v_i^2 \left( \prod_{m=1}^q g(\theta_{i,m}|d_{(m)}) \right)^2 \frac{g'_{k}(\theta_{i,j} | d_{(j)})g'_{t}(\theta_{i,s}|d_{(s)})}{g(\theta_{i,j} | d_{(j)})g(\theta_{i,s} | d_{(s)})} =
\end{equation}
\begin{equation}
=(r_i(d) - r_i^2(d)) \frac{g'_{k}(\theta_{i,j} | d_{(j)})g'_{t}(\theta_{i,s}|d_{(s)})}{g(\theta_{i,j} | d_j)g(\theta_{i,s} | d_{(s)})}.
\end{equation}
\fi
In case when both differentiation variables are the parameters of the same coordinate likelihood function for the same mixture component, the second derivative is:

\ificml
    \begin{equation}
    \frac{\partial^2}{\partial \theta_{i,j,k} \partial \theta_{i,j,t}} \log h(\theta|d) = 
    r_i(d) \frac{g''_{kt}(\theta_{i,j} | d_j)}{g(\theta_{i,j} | d_j )} - 
    \end{equation}
    $$
    -r_i(d)^2 \frac{g'_{k}(\theta_{i,j} | d_j)g'_{t}(\theta_{i,j} | d_j)}{(g(\theta_{i,j} | d_j))^2}.
    $$
\else
    \begin{equation}
    \frac{\partial^2}{\partial \theta_{i,j,k} \partial \theta_{i,j,t}} \log h(\theta|d) = 
    r_i(d) \frac{g''_{kt}(\theta_{i,j} | d_j)}{g(\theta_{i,j} | d_j )} - r_i(d)^2 \frac{g'_{k}(\theta_{i,j} | d_j)g'_{t}(\theta_{i,j} | d_j)}{(g(\theta_{i,j} | d_j))^2}.
    \end{equation}
\fi
In case when the differentiation variables are the parameters of different mixture components, the second derivative is:
\ificml
    \begin{equation}
    \frac{\partial^2}{\partial \theta_{i,j,k} \partial \theta_{u,s,t}} \log h(\theta|d) = -\frac{1}{h^2(\theta|d)} v_i \prod_{m=1}^q g(\theta_{i,m} | d_{(m)}) 
    \end{equation}
    \begin{equation}
    \frac{g'_{k}(\theta_{i,j}|d_{j})}{g(\theta_{i,j} | d_{(j)})} v_u \prod_{m=1}^q g(\theta_{u,m} | d_{(m)}) \frac{g'_{t}(\theta_{u,s}|d_{(s)})}{g(\theta_{u,s} | d_{(s)})} =
    \end{equation}
    \begin{equation}
    =-\frac{1}{h^2(\theta|d)} r_i(d) r_u(d)
    \frac{g'_{k}(\theta_{i,j}|d_{(j)})}{g(\theta_{i,j} | d_{(j)})}
    \frac{g'_{t}(\theta_{u,s}|d_{(s)})}{g(\theta_{u,s} | d_{(s)})}.
    \end{equation}
\else
    \begin{equation}
    \frac{\partial^2}{\partial \theta_{i,j,k} \partial \theta_{u,s,t}} \log h(\theta|d) = -\frac{1}{h^2(\theta|d)} v_i \prod_{m=1}^q g(\theta_{i,m} | d_{(m)}) \frac{g'_{k}(\theta_{i,j}|d_{j})}{g(\theta_{i,j} | d_{(j)})} \times 
    \end{equation}
    \begin{equation}
    \times
    v_u \prod_{m=1}^q g(\theta_{u,m} | d_{(m)}) \frac{g'_{t}(\theta_{u,s}|d_{(s)})}{g(\theta_{u,s} | d_{(s)})} = -\frac{1}{h^2(\theta|d)} r_i(d) r_u(d)
    \frac{g'_{k}(\theta_{i,j}|d_{(j)})}{g(\theta_{i,j} | d_{(j)})}
    \frac{g'_{t}(\theta_{u,s}|d_{(s)})}{g(\theta_{u,s} | d_{(s)})}.
    \end{equation}
\fi

{\bf Corollary.} In conditions of the Theorem, if for an observation responsibility of some mixture component $i$ is equal to $1$, then among the second derivatives, only the ones corresponding to the same mixture component and coordinate can differ from zero.

{\bf Proof.} Those derivatives which contain a term $r_i-r_i^2$ become zero, and the fact $r_i = 1$ leads to $r_u=0$ for $u \neq i$, so only the second case ($i=u, j=s$) of the theorem formulation can lead to a non-zero second order derivative.

The responsibility of an observation is equal to 1 (up to numerical precision) in some $95\%$ of cases in our experiments. Therefore second derivative matrices are very often sparse, having block-diagonal structure, which follows from the Corollary. It makes the meta-learning process for the Gaussian mixture models faster.

For the Gaussian mixture model with diagonal covariance matrices, the coordinate-wise function $g(\theta_{i,j} | d)$ and its derivatives w.r.t. the parameters $\theta_{i,j} = [\mu_{i,j}, \sigma_{i,j}]^T$ is given in the table \ref{tab:gmm_diff}.

\begin{table}
\centering
\begin{tabular}{|l|r|}
\hline
$g(\theta_{i,j} | d)$ & $(\sqrt{2 \pi} \sigma_{i,j})^{-1} e^{-\frac{(d_{(j)}-\mu_{i,j})^2}{2\sigma_{i,j}^2}}$ \\
\hline
$g'_{\mu}(\theta_{i,j} | d_{(j)}) $ & $\frac{d_{(j)}-\mu_{i,j}}{\sigma_{i,j}^2} g(\theta_{i,j} | d_{(j)})$ \\
\hline
$g'_{\sigma}(\theta_{i,j} | d_{(j)}) $ & 
$\sigma_{i,j}^{-1}(\frac{(d_{(j)}-\mu_{i,j})^2}{\sigma_{i,j}^2}-1)g(\theta_{i,j} | d_{(j)}) $\\
\hline
$g''_{\mu\mu} (\theta_{i,j} | d_{(j)})$ & 
$(-\frac{1}{\sigma_{i,j}^2} + \frac{(d_{(j)}-\mu_{i,j})^2}{\sigma_{i,j}^4}) g(\theta_{i,j} | d_{(j)})$ \\
\hline
$g''_{\sigma\sigma} (\theta_{i,j} | d_{(j)})$ & 
$\sigma_{i,j}^{-2} (2 - 5 \frac{(d_{(j)}-\mu_{i,j})^2}{\sigma_{i,j}^2} + \frac{(d_{(j)}-\mu_{i,j})^4}{\sigma_{i,j}^4}) g(\theta_{i,j} | d_{(j)})$ \\
\hline
$g''_{\sigma\mu}(\theta_{i,j} | d_{(j)})$ & 
$(-3\frac{d_{(j)}-\mu_{i,j}}{\sigma_{i,j}^3} + \frac{(d_{(j)}-\mu_{i,j})^3}{\sigma_{i,j}^5})
g(\theta_{i,j} | d_{(j)}) $ \\
\hline
\end{tabular}
\caption{Coordinate-wise likelihood function and its derivatives for the Gaussian mixture models.}\label{tab:gmm_diff}
\end{table}

\section{Acknowledgements}
Research was supported by the Russian MES grant RFMEFI61516X0003.
\fi

\ificml
\else
\section*{References}
\fi

\bibliography{refs}

\begin{thebibliography}{37}
\providecommand{\natexlab}[1]{#1}
\providecommand{\url}[1]{\texttt{#1}}
\providecommand{\href}[2]{#2}
\providecommand{\path}[1]{#1}
\providecommand{\eprint}[1]{\href{http://arxiv.org/abs/#1}{\path{#1}}}
\providecommand{\DOIprefix}{doi:}
\providecommand{\ArXivprefix}{arXiv:}
\providecommand{\URLprefix}{URL: }
\providecommand{\Pubmedprefix}{pmid:}
\providecommand{\doi}[1]{\href{http://dx.doi.org/#1}{\path{#1}}}
\providecommand{\Pubmed}[1]{\href{pmid:#1}{\path{#1}}}
\providecommand{\BIBand}{and}
\providecommand{\bibinfo}[2]{#2}
\ifx\xfnm\undefined \def\xfnm[#1]{\unskip,\space#1}\fi
\bibitem[{Tenenbaum et~al.(2011)Tenenbaum, Kemp, Griffiths and
  Goodman}]{tenenbaum2011grow}
\bibinfo{author}{Tenenbaum\xfnm[ J.B.]}, \bibinfo{author}{Kemp\xfnm[ C.]},
  \bibinfo{author}{Griffiths\xfnm[ T.L.]}, \bibinfo{author}{Goodman\xfnm[
  N.D.]}.
\newblock \bibinfo{title}{How to grow a mind: Statistics, structure, and
  abstraction}.
\newblock \bibinfo{journal}{Science}
  \bibinfo{year}{2011};\bibinfo{volume}{331}(\bibinfo{number}{6022}):\bibinfo{pages}{1279--1285}.
\bibitem[{Deng et~al.(2009)Deng, Dong, Socher, Li, Li and
  Fei-Fei}]{deng2009imagenet}
\bibinfo{author}{Deng\xfnm[ J.]}, \bibinfo{author}{Dong\xfnm[ W.]},
  \bibinfo{author}{Socher\xfnm[ R.]}, \bibinfo{author}{Li\xfnm[ L.J.]},
  \bibinfo{author}{Li\xfnm[ K.]}, \bibinfo{author}{Fei-Fei\xfnm[ L.]}.
\newblock \bibinfo{title}{Imagenet: A large-scale hierarchical image database}.
\newblock In: \bibinfo{booktitle}{Computer Vision and Pattern Recognition, IEEE
  Conference on}. \bibinfo{organization}{IEEE}; \bibinfo{year}{2009}, p.
  \bibinfo{pages}{248--255}.
\bibitem[{Guadarrama et~al.(2014)Guadarrama, Rodner, Saenko, Zhang, Farrell,
  Donahue et~al.}]{Guadarrama14}
\bibinfo{author}{Guadarrama\xfnm[ S.]}, \bibinfo{author}{Rodner\xfnm[ E.]},
  \bibinfo{author}{Saenko\xfnm[ K.]}, \bibinfo{author}{Zhang\xfnm[ N.]},
  \bibinfo{author}{Farrell\xfnm[ R.]}, \bibinfo{author}{Donahue\xfnm[ J.]},
  et~al.
\newblock \bibinfo{title}{Open-vocabulary object retrieval}.
\newblock In: \bibinfo{booktitle}{Robotics: science and systems};
  vol.~\bibinfo{volume}{2}. \bibinfo{year}{2014}, p.~\bibinfo{pages}{6}.
\bibitem[{Kumar and Seitz(2014)}]{Kumar14}
\bibinfo{author}{Kumar\xfnm[ N.]}, \bibinfo{author}{Seitz\xfnm[ S.]}.
\newblock \bibinfo{title}{Photo recall: Using the {Internet} to label your
  photos}.
\newblock In: \bibinfo{booktitle}{Proceedings of the IEEE Conference on
  Computer Vision and Pattern Recognition Workshops}. \bibinfo{year}{2014}, p.
  \bibinfo{pages}{771--778}.
\bibitem[{Tang et~al.(2010)Tang, Tappen, Sukthankar and
  Lampert}]{tang2010optimizing}
\bibinfo{author}{Tang\xfnm[ K.D.]}, \bibinfo{author}{Tappen\xfnm[ M.F.]},
  \bibinfo{author}{Sukthankar\xfnm[ R.]}, \bibinfo{author}{Lampert\xfnm[
  C.H.]}.
\newblock \bibinfo{title}{Optimizing one-shot recognition with micro-set
  learning}.
\newblock In: \bibinfo{booktitle}{Computer Vision and Pattern Recognition
  (CVPR), IEEE Conference on}. \bibinfo{organization}{IEEE};
  \bibinfo{year}{2010}, p. \bibinfo{pages}{3027--3034}.
\bibitem[{Vinyals et~al.(2016)Vinyals, Blundell, Lillicrap, Wierstra
  et~al.}]{vinyals2016matching}
\bibinfo{author}{Vinyals\xfnm[ O.]}, \bibinfo{author}{Blundell\xfnm[ C.]},
  \bibinfo{author}{Lillicrap\xfnm[ T.]}, \bibinfo{author}{Wierstra\xfnm[ D.]},
  et~al.
\newblock \bibinfo{title}{Matching networks for one shot learning}.
\newblock In: \bibinfo{booktitle}{Advances in Neural Information Processing
  Systems}. \bibinfo{year}{2016}, p. \bibinfo{pages}{3630--3638}.
\bibitem[{Wang and Hebert(2016)}]{wang2016learning}
\bibinfo{author}{Wang\xfnm[ Y.X.]}, \bibinfo{author}{Hebert\xfnm[ M.]}.
\newblock \bibinfo{title}{Learning to learn: Model regression networks for easy
  small sample learning}.
\newblock In: \bibinfo{booktitle}{European Conference on Computer Vision}.
  \bibinfo{organization}{Springer}; \bibinfo{year}{2016}, p.
  \bibinfo{pages}{616--634}.
\bibitem[{Ellis(1965)}]{ellis1965transfer}
\bibinfo{author}{Ellis\xfnm[ H.C.]}.
\newblock \bibinfo{title}{The transfer of learning.}
\newblock \bibinfo{publisher}{Macmillan}; \bibinfo{year}{1965}.
\bibitem[{Schmidhuber et~al.(1996)Schmidhuber, Zhao and
  Wiering}]{jurgen1996simple}
\bibinfo{author}{Schmidhuber\xfnm[ J.]}, \bibinfo{author}{Zhao\xfnm[ J.]},
  \bibinfo{author}{Wiering\xfnm[ M.]}.
\newblock \bibinfo{title}{Simple principles of metalearning}.
\newblock \bibinfo{type}{Tech. Rep.}; Istituto Dalle Molle Di Studi Sull
  Intelligenza Artificiale; \bibinfo{year}{1996}.
\bibitem[{Raina et~al.(2004)Raina, Shen, Ng and
  McCallum}]{raina2003classification}
\bibinfo{author}{Raina\xfnm[ R.]}, \bibinfo{author}{Shen\xfnm[ Y.]},
  \bibinfo{author}{Ng\xfnm[ A.Y.]}, \bibinfo{author}{McCallum\xfnm[ A.]}.
\newblock \bibinfo{title}{Classification with hybrid generative/discriminative
  models}.
\newblock In: \bibinfo{booktitle}{NIPS}; vol.~\bibinfo{volume}{16}.
  \bibinfo{year}{2004}, p. \bibinfo{pages}{545--552}.
\bibitem[{Holub and Perona(2005)}]{Holub05}
\bibinfo{author}{Holub\xfnm[ A.]}, \bibinfo{author}{Perona\xfnm[ P.]}.
\newblock \bibinfo{title}{A discriminative framework for modelling object
  classes}.
\newblock In: \bibinfo{booktitle}{{IEEE} Computer Society Conference on
  Computer Vision and Pattern Recognition {(CVPR})}. \bibinfo{year}{2005}, p.
  \bibinfo{pages}{664--671}.
\bibitem[{Lasserre et~al.(2006)Lasserre, Bishop and
  Minka}]{lasserre2006principled}
\bibinfo{author}{Lasserre\xfnm[ J.A.]}, \bibinfo{author}{Bishop\xfnm[ C.M.]},
  \bibinfo{author}{Minka\xfnm[ T.P.]}.
\newblock \bibinfo{title}{Principled hybrids of generative and discriminative
  models}.
\newblock In: \bibinfo{booktitle}{Computer Vision and Pattern Recognition, IEEE
  Conference on}; vol.~\bibinfo{volume}{1}. \bibinfo{organization}{IEEE};
  \bibinfo{year}{2006}, p. \bibinfo{pages}{87--94}.
\bibitem[{Minka(2005)}]{minka2005discriminative}
\bibinfo{author}{Minka\xfnm[ T.]}.
\newblock \bibinfo{title}{Discriminative models, not discriminative training}.
\newblock \bibinfo{type}{Tech. Rep.}; Technical Report MSR-TR-2005-144,
  Microsoft Research; \bibinfo{year}{2005}.
\bibitem[{Santoro et~al.(2016)Santoro, Bartunov, Botvinick, Wierstra and
  Lillicrap}]{santoro2016meta}
\bibinfo{author}{Santoro\xfnm[ A.]}, \bibinfo{author}{Bartunov\xfnm[ S.]},
  \bibinfo{author}{Botvinick\xfnm[ M.]}, \bibinfo{author}{Wierstra\xfnm[ D.]},
  \bibinfo{author}{Lillicrap\xfnm[ T.]}.
\newblock \bibinfo{title}{Meta-learning with memory-augmented neural networks}.
\newblock In: \bibinfo{booktitle}{Proceedings of The 33rd International
  Conference on Machine Learning}. \bibinfo{year}{2016}, p.
  \bibinfo{pages}{1842--1850}.
\bibitem[{Mensink et~al.(2013)Mensink, Verbeek, Perronnin and
  Csurka}]{Mensink13}
\bibinfo{author}{Mensink\xfnm[ T.]}, \bibinfo{author}{Verbeek\xfnm[ J.]},
  \bibinfo{author}{Perronnin\xfnm[ F.]}, \bibinfo{author}{Csurka\xfnm[ G.]}.
\newblock \bibinfo{title}{Distance-based image classification: Generalizing to
  new classes at near-zero cost}.
\newblock \bibinfo{journal}{IEEE Transactions on Pattern Analysis and Machine
  Intelligence}
  \bibinfo{year}{2013};\bibinfo{volume}{35}(\bibinfo{number}{11}):\bibinfo{pages}{2624--2637}.
\bibitem[{Fergus et~al.(2004)Fergus, Perona and Zisserman}]{fergus2004visual}
\bibinfo{author}{Fergus\xfnm[ R.]}, \bibinfo{author}{Perona\xfnm[ P.]},
  \bibinfo{author}{Zisserman\xfnm[ A.]}.
\newblock \bibinfo{title}{A visual category filter for {G}oogle images}.
\newblock In: \bibinfo{booktitle}{European Conference on Computer Vision}.
  \bibinfo{organization}{Springer}; \bibinfo{year}{2004}, p.
  \bibinfo{pages}{242--256}.
\bibitem[{Schroff et~al.(2011)Schroff, Criminisi and Zisserman}]{Schroff11}
\bibinfo{author}{Schroff\xfnm[ F.]}, \bibinfo{author}{Criminisi\xfnm[ A.]},
  \bibinfo{author}{Zisserman\xfnm[ A.]}.
\newblock \bibinfo{title}{Harvesting image databases from the web}.
\newblock \bibinfo{journal}{IEEE Transactions on Pattern Analysis and Machine
  Intelligence}
  \bibinfo{year}{2011};\bibinfo{volume}{33}(\bibinfo{number}{4}):\bibinfo{pages}{754--766}.
\bibitem[{Arandjelovic and Zisserman(2012)}]{Arandjelovic12}
\bibinfo{author}{Arandjelovic\xfnm[ R.]}, \bibinfo{author}{Zisserman\xfnm[
  A.]}.
\newblock \bibinfo{title}{Multiple queries for large scale specific object
  retrieval.}
\newblock In: \bibinfo{booktitle}{BMVC}. \bibinfo{year}{2012}, p.
  \bibinfo{pages}{1--11}.
\bibitem[{Chatfield and Zisserman(2012)}]{Chatfield12}
\bibinfo{author}{Chatfield\xfnm[ K.]}, \bibinfo{author}{Zisserman\xfnm[ A.]}.
\newblock \bibinfo{title}{Visor: Towards on-the-fly large-scale object category
  retrieval}.
\newblock In: \bibinfo{booktitle}{Asian Conference on Computer Vision}.
  \bibinfo{organization}{Springer}; \bibinfo{year}{2012}, p.
  \bibinfo{pages}{432--446}.
\bibitem[{Chatfield et~al.(2015)Chatfield, Arandjelovi{\'c}, Parkhi and
  Zisserman}]{Chatfield15}
\bibinfo{author}{Chatfield\xfnm[ K.]}, \bibinfo{author}{Arandjelovi{\'c}\xfnm[
  R.]}, \bibinfo{author}{Parkhi\xfnm[ O.]}, \bibinfo{author}{Zisserman\xfnm[
  A.]}.
\newblock \bibinfo{title}{On-the-fly learning for visual search of large-scale
  image and video datasets}.
\newblock \bibinfo{journal}{International Journal of Multimedia Information
  Retrieval}
  \bibinfo{year}{2015};\bibinfo{volume}{4}(\bibinfo{number}{2}):\bibinfo{pages}{75--93}.
\bibitem[{Tenorth et~al.(2011)Tenorth, Klank, Pangercic and Beetz}]{Tenorth11}
\bibinfo{author}{Tenorth\xfnm[ M.]}, \bibinfo{author}{Klank\xfnm[ U.]},
  \bibinfo{author}{Pangercic\xfnm[ D.]}, \bibinfo{author}{Beetz\xfnm[ M.]}.
\newblock \bibinfo{title}{Web-enabled robots}.
\newblock \bibinfo{journal}{Robotics Automation Magazine, IEEE}
  \bibinfo{year}{2011};\bibinfo{volume}{18}(\bibinfo{number}{2}):\bibinfo{pages}{58--68}.
\bibitem[{Lai and Fox(2010)}]{Lai10}
\bibinfo{author}{Lai\xfnm[ K.]}, \bibinfo{author}{Fox\xfnm[ D.]}.
\newblock \bibinfo{title}{Object recognition in 3d point clouds using web data
  and domain adaptation}.
\newblock \bibinfo{journal}{I J Robotic Res}
  \bibinfo{year}{2010};\bibinfo{volume}{29}(\bibinfo{number}{8}):\bibinfo{pages}{1019--1037}.
\bibitem[{Jain and Varma(2011)}]{jain2011learning}
\bibinfo{author}{Jain\xfnm[ V.]}, \bibinfo{author}{Varma\xfnm[ M.]}.
\newblock \bibinfo{title}{Learning to re-rank: query-dependent image re-ranking
  using click data}.
\newblock In: \bibinfo{booktitle}{Proceedings of the 20th international
  conference on World wide web}. \bibinfo{organization}{ACM};
  \bibinfo{year}{2011}, p. \bibinfo{pages}{277--286}.
\bibitem[{Yang and Hanjalic(2010)}]{yang2010supervised}
\bibinfo{author}{Yang\xfnm[ L.]}, \bibinfo{author}{Hanjalic\xfnm[ A.]}.
\newblock \bibinfo{title}{Supervised reranking for web image search}.
\newblock In: \bibinfo{booktitle}{Proceedings of the 18th ACM international
  conference on Multimedia}. \bibinfo{organization}{ACM}; \bibinfo{year}{2010},
  p. \bibinfo{pages}{183--192}.
\bibitem[{Schwarz et~al.(1978)}]{schwarz1978estimating}
\bibinfo{author}{Schwarz\xfnm[ G.]}, et~al.
\newblock \bibinfo{title}{Estimating the dimension of a model}.
\newblock \bibinfo{journal}{The annals of statistics}
  \bibinfo{year}{1978};\bibinfo{volume}{6}(\bibinfo{number}{2}):\bibinfo{pages}{461--464}.
\bibitem[{Ustinova and Lempitsky(2016)}]{ustinova2016learning}
\bibinfo{author}{Ustinova\xfnm[ E.]}, \bibinfo{author}{Lempitsky\xfnm[ V.]}.
\newblock \bibinfo{title}{Learning deep embeddings with histogram loss}.
\newblock In: \bibinfo{booktitle}{Advances in Neural Information Processing
  Systems (NIPS)}. \bibinfo{year}{2016}, p. \bibinfo{pages}{4170--4178}.
\bibitem[{Kingma and Ba(2014)}]{kingma2014adam}
\bibinfo{author}{Kingma\xfnm[ D.]}, \bibinfo{author}{Ba\xfnm[ J.]}.
\newblock \bibinfo{title}{Adam: A method for stochastic optimization}.
\newblock \bibinfo{journal}{arXiv preprint arXiv:14126980}
  \bibinfo{year}{2014};.
\bibitem[{Jia et~al.(2014)Jia, Shelhamer, Donahue, Karayev, Long, Girshick
  et~al.}]{jia2014caffe}
\bibinfo{author}{Jia\xfnm[ Y.]}, \bibinfo{author}{Shelhamer\xfnm[ E.]},
  \bibinfo{author}{Donahue\xfnm[ J.]}, \bibinfo{author}{Karayev\xfnm[ S.]},
  \bibinfo{author}{Long\xfnm[ J.]}, \bibinfo{author}{Girshick\xfnm[ R.]},
  et~al.
\newblock \bibinfo{title}{Caffe: Convolutional architecture for fast feature
  embedding}.
\newblock In: \bibinfo{booktitle}{Proceedings of the 22nd ACM International
  Conference on Multimedia}. \bibinfo{organization}{ACM}; \bibinfo{year}{2014},
  p. \bibinfo{pages}{675--678}.
\bibitem[{Krizhevsky et~al.(2012)Krizhevsky, Sutskever and
  Hinton}]{krizhevsky2012imagenet}
\bibinfo{author}{Krizhevsky\xfnm[ A.]}, \bibinfo{author}{Sutskever\xfnm[ I.]},
  \bibinfo{author}{Hinton\xfnm[ G.E.]}.
\newblock \bibinfo{title}{Imagenet classification with deep convolutional
  neural networks}.
\newblock In: \bibinfo{booktitle}{Advances in neural information processing
  systems}. \bibinfo{year}{2012}, p. \bibinfo{pages}{1097--1105}.
\bibitem[{LeCun et~al.(1998)LeCun, Bottou, Bengio and
  Haffner}]{lecun1998gradient}
\bibinfo{author}{LeCun\xfnm[ Y.]}, \bibinfo{author}{Bottou\xfnm[ L.]},
  \bibinfo{author}{Bengio\xfnm[ Y.]}, \bibinfo{author}{Haffner\xfnm[ P.]}.
\newblock \bibinfo{title}{Gradient-based learning applied to document
  recognition}.
\newblock \bibinfo{journal}{Proceedings of the IEEE}
  \bibinfo{year}{1998};\bibinfo{volume}{86}(\bibinfo{number}{11}):\bibinfo{pages}{2278--2324}.
\bibitem[{Bergstra et~al.(2010)Bergstra, Breuleux, Bastien, Lamblin, Pascanu,
  Desjardins et~al.}]{bergstra2010theano}
\bibinfo{author}{Bergstra\xfnm[ J.]}, \bibinfo{author}{Breuleux\xfnm[ O.]},
  \bibinfo{author}{Bastien\xfnm[ F.]}, \bibinfo{author}{Lamblin\xfnm[ P.]},
  \bibinfo{author}{Pascanu\xfnm[ R.]}, \bibinfo{author}{Desjardins\xfnm[ G.]},
  et~al.
\newblock \bibinfo{title}{Theano: A {CPU} and {GPU} math compiler in python}.
\newblock In: \bibinfo{booktitle}{Proc. 9th Python in Science Conf}.
  \bibinfo{year}{2010}, p. \bibinfo{pages}{1--7}.
\bibitem[{Koch et~al.(2015)Koch, Zemel and Salakhutdinov}]{koch2015siamese}
\bibinfo{author}{Koch\xfnm[ G.]}, \bibinfo{author}{Zemel\xfnm[ R.]},
  \bibinfo{author}{Salakhutdinov\xfnm[ R.]}.
\newblock \bibinfo{title}{Siamese neural networks for one-shot image
  recognition}.
\newblock In: \bibinfo{booktitle}{ICML Deep Learning Workshop}.
  \bibinfo{year}{2015},.
\bibitem[{Russakovsky et~al.(2015)Russakovsky, Deng, Su, Krause, Satheesh, Ma
  et~al.}]{ILSVRC15}
\bibinfo{author}{Russakovsky\xfnm[ O.]}, \bibinfo{author}{Deng\xfnm[ J.]},
  \bibinfo{author}{Su\xfnm[ H.]}, \bibinfo{author}{Krause\xfnm[ J.]},
  \bibinfo{author}{Satheesh\xfnm[ S.]}, \bibinfo{author}{Ma\xfnm[ S.]}, et~al.
\newblock \bibinfo{title}{{ImageNet Large Scale Visual Recognition Challenge}}.
\newblock \bibinfo{journal}{International Journal of Computer Vision (IJCV)}
  \bibinfo{year}{2015};\bibinfo{volume}{115}(\bibinfo{number}{3}):\bibinfo{pages}{211--252}.
\newblock \DOIprefix\doi{10.1007/s11263-015-0816-y}.
\bibitem[{Lai et~al.(2011)Lai, Bo, Ren and Fox}]{lai2011large}
\bibinfo{author}{Lai\xfnm[ K.]}, \bibinfo{author}{Bo\xfnm[ L.]},
  \bibinfo{author}{Ren\xfnm[ X.]}, \bibinfo{author}{Fox\xfnm[ D.]}.
\newblock \bibinfo{title}{A large-scale hierarchical multi-view {RGB-D} object
  dataset}.
\newblock In: \bibinfo{booktitle}{Robotics and Automation (ICRA), 2011 IEEE
  International Conference on}. \bibinfo{organization}{IEEE};
  \bibinfo{year}{2011}, p. \bibinfo{pages}{1817--1824}.
\bibitem[{Lai et~al.(2014)Lai, Bo and Fox}]{lai2014unsupervised}
\bibinfo{author}{Lai\xfnm[ K.]}, \bibinfo{author}{Bo\xfnm[ L.]},
  \bibinfo{author}{Fox\xfnm[ D.]}.
\newblock \bibinfo{title}{Unsupervised feature learning for 3d scene labeling}.
\newblock In: \bibinfo{booktitle}{Robotics and Automation (ICRA), IEEE
  International Conference on}. \bibinfo{organization}{IEEE};
  \bibinfo{year}{2014}, p. \bibinfo{pages}{3050--3057}.
\bibitem[{Nilsback and Zisserman(2008)}]{nilsback2008automated}
\bibinfo{author}{Nilsback\xfnm[ M.E.]}, \bibinfo{author}{Zisserman\xfnm[ A.]}.
\newblock \bibinfo{title}{Automated flower classification over a large number
  of classes}.
\newblock In: \bibinfo{booktitle}{Computer Vision, Graphics \& Image
  Processing, Sixth Indian Conference on}. \bibinfo{organization}{IEEE};
  \bibinfo{year}{2008}, p. \bibinfo{pages}{722--729}.
\bibitem[{Lake et~al.(2011)Lake, Salakhutdinov, Gross and
  Tenenbaum}]{lake2011one}
\bibinfo{author}{Lake\xfnm[ B.M.]}, \bibinfo{author}{Salakhutdinov\xfnm[ R.]},
  \bibinfo{author}{Gross\xfnm[ J.]}, \bibinfo{author}{Tenenbaum\xfnm[ J.B.]}.
\newblock \bibinfo{title}{One shot learning of simple visual concepts}.
\newblock In: \bibinfo{booktitle}{CogSci}; vol. \bibinfo{volume}{172}.
  \bibinfo{year}{2011}, p.~\bibinfo{pages}{2}.

\end{thebibliography}

\end{document}